\documentclass[times,twocolumn,final,authoryear]{elsarticle}

\usepackage{ycviu}
\usepackage{framed,multirow}

\usepackage{amssymb}
\usepackage{amsmath}
\usepackage{latexsym}
\usepackage{graphicx}
\usepackage{booktabs}
\usepackage{algorithm} 
\usepackage{algorithmic} 
\usepackage{textcomp}

\newcommand{\revise}[1]{{\color{black}{#1}}}

\usepackage{url}
\usepackage{xcolor}
\usepackage{xspace}
\usepackage{subcaption}
\usepackage{makecell}

\usepackage[capitalize]{cleveref}
\crefname{section}{Sec.}{Secs.}
\Crefname{section}{Section}{Sections}
\Crefname{table}{Table}{Tables}
\crefname{table}{Tab.}{Tabs.}

\makeatletter
\def\onedot{.\null\xspace}

\def\eg{\emph{e.g}\onedot} 
\def\ie{\emph{i.e}\onedot}

\def\etal{\emph{et al}\onedot}
\makeatother

\definecolor{newcolor}{rgb}{.8,.349,.1}

\journal{Computer Vision and Image Understanding}

\begin{document}

\ifpreprint
  \setcounter{page}{1}
\else
  \setcounter{page}{1}
\fi

\begin{frontmatter}

\title{To Make Yourself Invisible with Adversarial Semantic Contours}

\author[1]{Yichi \snm{Zhang}\fnref{contrib}}
\fntext[contrib]{Equal Contribution}
\ead{zyc22@mails.tsinghua.edu.cn}
\author[2]{Zijian \snm{Zhu}\fnref{contrib}}
\author[1]{Hang \snm{Su}}
\author[1]{Jun \snm{Zhu}\corref{cor1}}
\cortext[cor1]{Corresponding author}
\ead{dcszj@mail.tsinghua.edu.cn}
\author[2]{Shibao \snm{Zheng}}
\author[3]{Yuan \snm{He}}
\author[3]{Hui \snm{Xue}}

\address[1]{Department of Computer Science and Technology, Institute for Artificial Intelligence, THBI Lab, Tsinghua University, Beijing 100084, China}
\address[2]{Institute of Image Communication and Network Engineering, Shanghai Jiao Tong University, Shanghai 200240, China}
\address[3]{Alibaba Group, Hangzhou 311121, China}

\received{1 May 2013}
\finalform{10 May 2013}
\accepted{13 May 2013}
\availableonline{15 May 2013}
\communicated{S. Sarkar}

\begin{abstract}
Modern object detectors are vulnerable to adversarial examples, which may bring risks to real-world applications. The sparse attack 
is an important
task which, compared with the popular adversarial perturbation on the whole image, needs to select the potential pixels that is generally regularized by an $\ell_0$-norm constraint, 
and simultaneously optimize the corresponding texture. The non-differentiability of $\ell_0$ norm brings challenges and many works on attacking object detection adopted manually-designed patterns to address them, which are meaningless and independent of objects, and therefore lead to relatively poor attack performance.
 In this paper, we propose Adversarial Semantic Contour (ASC), an MAP estimate of a Bayesian formulation of sparse attack with a deceived prior of object contour. The object contour prior effectively reduces the search space of pixel selection and improves the attack by introducing more semantic bias. Extensive experiments demonstrate that ASC can corrupt the prediction of 9 modern detectors with different architectures (\eg, one-stage, two-stage and Transformer) by modifying fewer than 5\% of the pixels of the object area in COCO in white-box scenario and around 10\% of those in black-box scenario. We further extend the attack to datasets for autonomous driving systems to verify the effectiveness. We conclude with cautions about contour being the common weakness of object detectors with various architecture and the care needed in applying them in safety-sensitive scenarios.
\end{abstract}


\end{frontmatter}

\section{Introduction}
\label{sec:intro}

Deep Neural Networks (DNNs) have obtained remarkable success on object detection~\citep{ren2015faster,he2017mask,Liu_2016ssd,yolov3,carion2020detr}, which involves simultaneous object localization and classification. Such methods have been widely used in security-sensitive scenarios, including video surveillance, facial recognition and autonomous driving systems. 
However, recent research has shown the vulnerability of DNN-based models~\citep{szegedy2013intriguing,goodfellow2014explaining,madry2017towards,dong2018boosting}, and modern detectors are vulnerable to carefully designed adversarial perturbations~\citep{carlini2017adversarial,lu2017adversarial,xie2017adversarial,shen2019advspade,wu2020dpattack}, which raises safety concerns in real-world applications. Therefore, it is crucial to investigate the adversarial weakness of object detection, which can then be addressed to strengthen the robustness of object detectors and further improve safety in scenarios such as auto-driving.

\begin{figure}[t]
\centering
\begin{minipage}[]{.32\linewidth}
{               
\includegraphics[width=2.8cm]{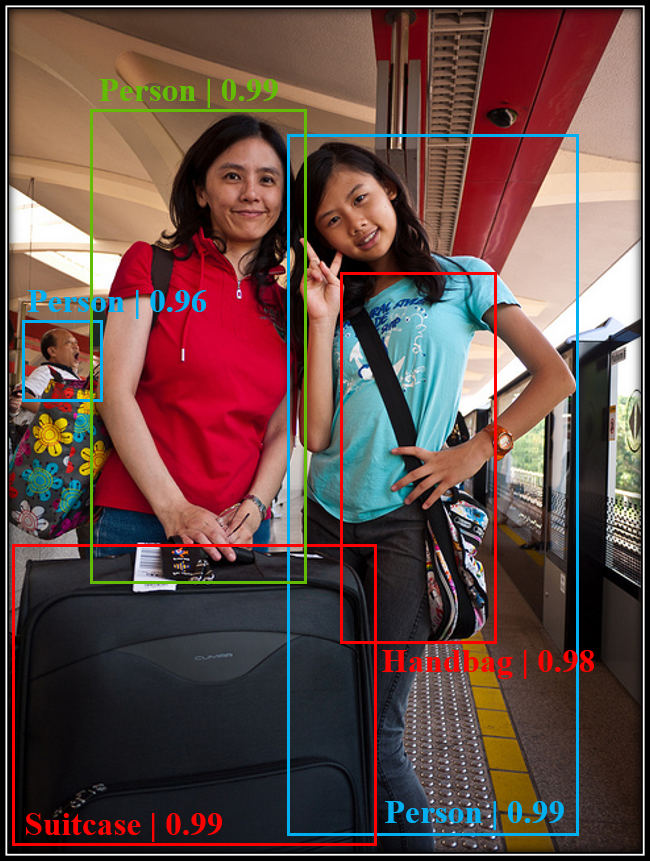}}
\subcaption{Object Detection}
\end{minipage}
\hspace{0in}
\begin{minipage}[]{.32\linewidth}
{
\includegraphics[width=2.8cm]{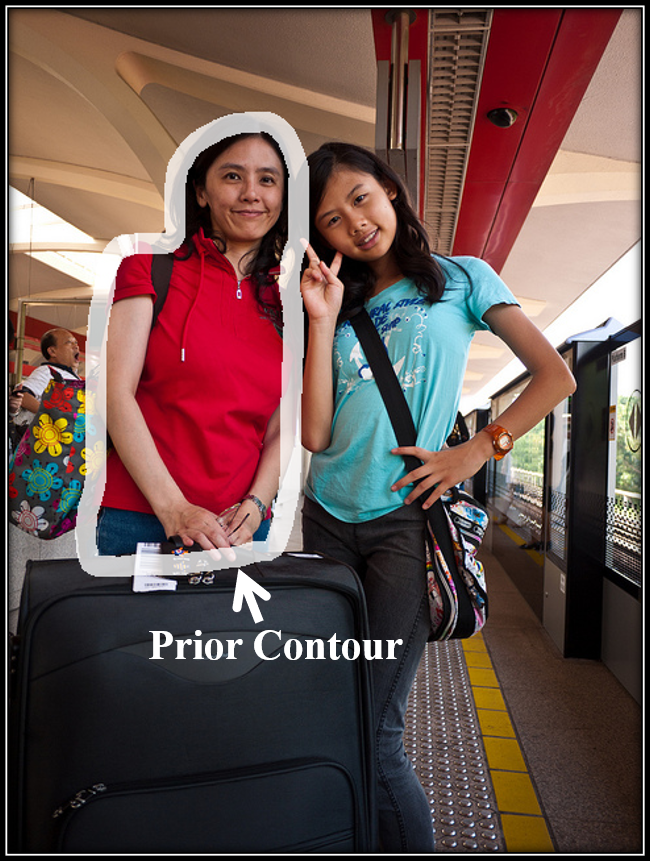}}
\subcaption{Prior Contour}
\end{minipage}
\hspace{0in}
\begin{minipage}[]{.32\linewidth}
{               
\includegraphics[width=2.8cm]{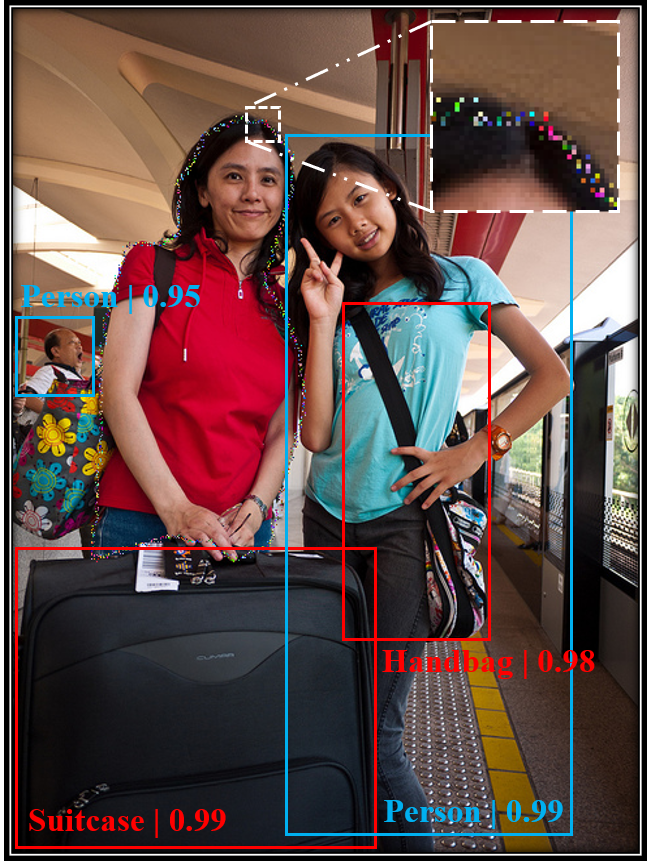}}
\subcaption{Invisible with ASC}
\end{minipage}
\caption{We conduct a sparse attack on object detectors guided by a contour prior, which can be rough and outlined manually. With pixel sampling and texture optimization, we successfully make the woman in the green bounding box invisible to DETR~\citep{carion2020detr}, while other objects remain detectable, by modifying only 2\% of the pixels in the object area.
\vspace{-4ex}
}
\label{pic:example}
\end{figure}

Methods for object detection have much more diverse pipelines compared to those for image classification, which makes it nontrivial to conduct adversarial attacks within a generic framework. Specifically, modern detectors can be classified into one-stage detectors (\eg, Yolo~\citep{redmon2016you,redmon2016yolo9000,yolov3}), which use a single forward pass to make predictions, and two-stage detectors (\eg, R-CNN~\citep{girshick2014rich}, Faster R-CNN\citep{ren2015faster}), which generate region proposals by Region Proposal Network (RPN) and conduct classification on them. RPN brings with redundant region proposals and smaller feature maps of Region of Interest (RoI), which may neutralize adversarial perturbations and lead to more challenges for adversarial attack on two-stage detectors. The rising Transformer architecture~\citep{parmar2018image,guo2021pct,liu2021swin} is being applied to object detectors, either as backbone~\citep{liu2021swin} or as detection head (\eg, DETR and its incremental versions~\citep{carion2020detr,zhu2020deformable, liu2022dabdetr}). Some studies reveal that it's more difficult to conduct adversarial attacks on Vision Transformers (ViT), because they learn more generalized contextual information~\citep{shao2021adversarial}, but the adversarial robustness of Transformer-based detectors still remain rarely studied.

There has been extensive research on generating adversarial examples for object detection. 
Early methods like DAG~\citep{xie2017adversarial} and UEA~\citep{wei2018transferable} perturb the whole digital image with imperceptible noise bounded by $\ell_\infty$ or $\ell_2$ norm. Recently, sparse attack has drawn an increasing amount of attention~\citep{liu2018dpatch,wu2020dpattack,thys2019fooling,advpatch,huang2020universal}. \revise{It explores the robustness of neural network from a different setting, where a constraint on $\ell_0$ norm of the perturbation, \ie, the number of pixels to be modified, is imposed, instead of limiting the perturbation budget. It's equally worthy as global attack to be studied in both algorithms and applications, because it's a typical NP-hard problem and its setting is closer to that in real-world attack, where only limited area of the environment can be disturbed.}

However, this is a problem of constrained optimization and also technically more challenging to solve as it requires to optimize on the texture and select attacked pixels simultaneously, and the pixel selection part is generally a combinatorial problem that is sparse and non-differentiable.  
Therefore, we have to resort to find approximate solutions. Traditional $\ell_0$ attack methods in image classification~\citep{carlini2017towards,croce2019sparse} usually disturb scattered pixels over the whole image, which are selected with heuristic or other strategies. Apart from those, many recent studies on sparse attack on object detectors focus more on perturbing areas which meet certain connected patterns that are commonly manually predefined and independent of the object properties~\citep{huang2020universal,saha2019adversarial,thys2019fooling,liu2018dpatch,wu2020dpattack,bao2020sparse}. This is also known as patch-based attack. From another perspective, these predesigned patterns serve as prior knowledge for pixel selection. However, the meaningless shape and location of the prior patterns are not able to fully capture the object characteristics and can sacrifice the effectiveness in terms of the attack success rate or the $\ell_0$ norm and the efficiency of texture optimization.

To address the discussed problem, we present Adversarial Semantic Contour (ASC) to improve the sparse attack performance on object detectors. 
We define ASC as an Maximum A Posteriori (MAP) estimate under a Bayesian formulation of $\ell_0$-constrained sparse attack and design a prior over the object contours for pixel selection. ASC can effectively attack various object detectors in an efficient manner. 
In particular, our prior knowledge of object contour comes from the insights of previous works that reveal the importance of contour and shape for object detection. In~\citep{alexe2010object}, the authors thoroughly analyze the question ``What is an object?'', 
and point out that an object should have closed boundaries and a different appearance against the background. This implies that the object contours carry significant information about the object. Additionally, it has shown that object detectors may benefit from shape cues~\citep{geirhos2018imagenet}.
Based on the informative prior contour, we can optimize the texture with a gradient method similar to PGD~\citep{madry2017towards} technically. To further uplift the attack performance, we model a prior distribution defined by the contour and adopt MAP estimation to select the pixels to be perturbed via sampling. The introduction of contour fits the characteristics of the task of object detection, and significantly enhance the attack.

We conduct comprehensive experiments to verify the attack effectiveness and efficiency of our method. We first apply ASC in the popular dataset COCO~\citep{lin2014microsoft} and in auto-driving with two different datasets of CityScapes~\citep{Cordts2016Cityscapes} and BDD100K~\citep{bdd100k} under the task of object vanishing. Experimental results show that, with the same $\ell_0$ norm constraints (\eg, no more than 5\% of the object area in the COCO case), we can achieve better attack both than the popular predesigned patterns and than the traditional $\ell_0$ attack methods with further optimization in pixel selection. The convergence curves and transfer-based attacks indicate that ASC achieves a successful attack with fewer iterations of texture optimization and better transferability in black-box adversarial attacks thanks to the appropriate contour priors. Additional studies also prove the power of our methods in different tasks and on defense models. Our results suggest that semantic contour is a common weakness to various detectors by testing on 9 different modern detectors.

In summary, our contributions are:
\begin{itemize}
    \item We introduce a simple yet effective prior of object contours by considering the characteristics of an object detector, which promotes attack effectiveness and efficiency along with transferability as a consequence.
    \item We propose a novel Bayesian framework for adversarial attacks on object detection, which provides a generic solution for critical sparse attacks on object detection with $\ell_0$ constraints.
    \item Comprehensive experiments are conducted on nine modern detectors (including one-stage, two-stage detectors and Detection Transformers) within COCO and two auto-driving datasets, and the results verify the effectiveness, efficiency and transferability of our method, which reveal the common weakness of object detectors and indicate the potential safety hazards of object detection and its application, \eg, auto-driving systems.
\end{itemize}

\section{Related Work}
\label{sec:related}

In this section, we briefly review previous methods for object detection and adversarial attacks on that.

\subsection{Object Detection}
\label{sec:object}

In ``What is an object?''~\citep{alexe2010object}, three distinctive characteristics of objects are investigated, including ``a closed boundary in space'', ``a different appearance'' and ``salience against the background''. Additionally, four cues for objectness measurement are applied. Multi-Scale Saliency and Color Contrast are closely related to foreground color, while Edge Density and Straddleness correlate more with object contour. An object can be detected by optimizing the parameters of the measurement in a Bayesian framework.
Taking the parameter $\phi$ of Edge Detection for instance, the optimal $\phi^*$ is maximized via the posterior probability of windows covering the ground-truth labels as positives as
\begin{align}
\label{eq:ed}
    \phi^*&=\mathop{\arg\max}_\phi\prod_{w\in\mathcal{W}^{\text{obj}}}p_\phi(\text{obj}|\text{ED}(w,\phi)) \\
    &=\mathop{\arg\max}_\phi\prod_{w\in\mathcal{W}^{\text{obj}}}\frac{p_\phi(\text{ED}(w,\phi)|\text{obj})\cdot p(\text{obj})}{\sum_{c\in\{\text{obj},\text{bg}\}}p_\phi(\text{ED}(w,\phi)|c)\cdot p(c)}, \nonumber
\end{align}
where $\text{ED}(w,\phi)$ is the edge density of window $w$ with parameter $\phi$, $p(\text{obj})$ is the probability of the window to be predicted as an object and $\mathcal{W}^{\text{obj}}$ is the set of windows bounding a ground-truth object.

Modern DNN-based object detectors are mainly categorized into one-stage detectors, such as Yolo \citep{redmon2016you,redmon2016yolo9000,yolov3} and SSD \citep{Liu_2016ssd}, which make predictions with one forward pass, and two-stage detectors, such as Faster R-CNN~\citep{ren2015faster} and Mask R-CNN~\citep{he2017mask}, which are mostly based on R-CNN~\citep{girshick2014rich}. In addition to ResNet~\citep{he2016deep} and VGGNet~\citep{simonyan2014very}, recent Vision Transformers (ViT) like Swin-T~\citep{liu2021swin} can also be used as backbone network to extract features. Besides, Detection Transformer (DETR)~\citep{carion2020detr} applies Transformer~\citep{vaswani2017attention} architecture and removes post-processing operations such as NMS. The adversarial robustness of ViTs compared to CNNs is under heated discussion~\citep{mahmood2021robustness,bai2021transformers,shao2021adversarial} and that of Transformer-based detectors still remains further study.

\subsection{Adversarial Attack on Object Detection}

Adversarial examples for DNNs were first put forward in~\citep{szegedy2013intriguing}. DNN-based object detectors have been proven to be vulnerable to adversarial examples~\citep{carlini2017adversarial,lu2017adversarial,xie2017adversarial, wei2018transferable}. The generation of adversarial perturbations is often a constrained optimization problem and existing methods can be categorized according to their $\ell_p$ constraints. Some methods apply $\ell_\infty$ as the bound for perturbations to create noise imperceptible to humans, such as DAG~\citep{xie2017adversarial} and UEA~\citep{wei2018transferable}. Nevertheless, these methods add perturbations over the whole image, which can be impractical in a real-world attack. Compared to the $\ell_\infty$ attack, the $\ell_0$ attack, also known as sparse attack, which aims to modify a limited number of pixels, is also important to be studied. 

Sparse attack needs to select the attacked pixels and optimize the texture simultaneously, while the pixel selection is non-differentiable. In the work of C\&W~\citep{carlini2017towards}, an $\ell_0$ attack pipeline using an $\ell_2$ adversary is proposed, whose pixel selection tightly relies on target models and takes many rounds of attacks to get the solution. PGD$_0$~\citep{croce2019sparse} selects a given number of pixels to modify in every iteration of texture optimization. However, these methods are proposed in the scenario of classification and always modified scattered pixels. \citep{advpatch} first introduces the adversarial patch, which has become popular for object detection~\citep{liu2018dpatch,thys2019fooling,saha2019adversarial,wu2020dpattack}. Manually designed patterns, such as square patches~\citep{liu2018dpatch,thys2019fooling,saha2019adversarial} and grids~\citep{wu2020dpattack}, are adopted to avoid time-costly search and need only iterative optimization for their textures. However, the meaningless patterns, which are pre-designed with certain shapes irrelevant to the object characteristics, limit the attack effectiveness in terms of the attack success rate or the $\ell_0$ cost and the convergence efficiency for texture optimization.

\section{Methodology}
\label{sec:method}

In this section, we present our formulation of sparse attack on an object detector constrained by $\ell_0$ norm within a Bayesian framework and introduce a prior of object contour for the pixel selection.

In general, to attack with $\ell_0$ constraints, we need to decide which pixels are to be disturbed and their resulting adversarial textures. We consider a formulation of finding the optimal adversarial noise for each image in a general form as
\begin{equation}
\begin{split}
    \delta^* = \mathop{\arg\max}_{\delta}J(g_\omega(x+ \delta),y), \;
    \operatorname{ s.t. }\;\ell_0(\delta)\leq N_0,\\
\end{split}
    \label{eq:general}
\end{equation}
where $x$ is the original image, $y$ is its corresponding ground-truth label, $g_\omega$ denotes the detector with parameter $\omega$ representing the structure and weights of the model, $J(g_\omega(\cdot),y)$ denotes the objective function that describes the similarity between the detector output and the ground-truth label $y$, and $\delta$ is the adversarial noise. We maximize the objective function under the constraint that the $\ell_0$ norm of $\delta$ is smaller than a constant $N_0$. Based on \cref{eq:ed}, the detector prediction is a likelihood given the parameters and input. In this perspective, we can rewrite our formulation of an $\ell_0$ attack as an Maximum Likelihood estimation (MLE) with an $\ell_0$ constraint:
\begin{equation}
    \delta^* = \mathop{\arg\max}_{\delta}\log p_\omega(\bar{y}|x+\delta), \;
    \operatorname{ s.t. }\;\ell_0(\delta)\leq N_0,
    \label{eq:likelihood}
\end{equation}
where we substitute maximizing the objective function $J$ with maximizing the log-likelihood of the wrong prediction of $\bar{y}$.

One challenge of problem (\ref{eq:likelihood}) is that $\delta$ involves pixel selection and texture optimization simultaneously, which are coupled with each other. We propose to divide it into two parts as $\delta=t\odot M$ and alternately optimize them. Here, $t\in\mathbb{R}^{h\times w\times 3}$ is the texture, $M\in \{0,1\}^{h\times w}$ is the binary mask for pixel selection and $\odot$ is element-wise multiplication. Then, the optimization of $\delta$ can be decoupled as
\begin{equation}
\begin{split}
    &\max_{M,t}\log p_\omega(\bar{y}|x+t\odot M)\\
    =&\max_{M,t}\log p_\omega(\bar{y}|M,t),\;\operatorname{ s.t. }\;\ell_0(M)\leq N_0,
\end{split}
\label{eq:reform}
\end{equation}
where the $\ell_0$ constraint only concerns $M$ and we ignore the fixed input $x$ in the condition of likelihood for notational clarity.

The challenge of this problem is that the optimization of $M$ with an $\ell_0$ constraint is non-differentiable and NP-hard in general. We may get a good approximation of the $\ell_0$ norm using $\ell_1$ norm, but the non-differentiable modules in diverse object detectors~\citep{girshick2014rich} still make the problem intractable only with gradient methods. We here adopt appropriate prior knowledge to better approximate the optimization, motivated by the claim in~\citep{a13010008randomize} that introducing non-uniform randomization priors can cope effectively with non-smooth optimization. We introduce a prior distribution for $\delta = (M,t)$ which can guide the optimization in a Bayesian framework with MAP estimation constrained by $\ell_0$ norm as 
\begin{equation}
\begin{split}
    M^*,t^*&=\mathop{\arg\max}\limits_{M,t} \log\big( p_\omega(\bar{y}|M,t)p(M,t)\big)\\
    &=\mathop{\arg\max}\limits_{M,t} \log\big(p_\omega(\bar{y}|M,t) p(t|M)p(M)\big),\\
    &\operatorname{ s.t. }\;\ell_0(M)\leq N_0,
\end{split}
\label{eq:bayesian}
\end{equation}
where $p(M,t)$ is the joint prior distribution for $\delta=(M,t)$ based on prior knowledge, and it can be factorized as $p(t|M)p(M)$. 

The above formulation subsumes some existing methods that solve~\cref{eq:likelihood}. For instance, the methods using heuristic strategies (such as C\&W-$\ell_0$~\citep{carlini2017towards} and PGD$_0$\citep{croce2019sparse}) have $p(M,t)$ as a uniform distribution. As for sparse patch-based methods, such as fixed pre-designed patterns like squares and grids~\citep{liu2018dpatch,thys2019fooling,wu2020dpattack}, their prior distribution for $p(M)$ can be seen as an impulse function and they set $p(t|M)$ uniformly with no special assumptions because texture $t$ can be optimized based on gradient methods given a certain binary mask $M$. However, these patterns lack object characteristics and information, and cannot achieve successful attacks effectively on various detectors. Similarly, we also focus on $p(M)$ in our work but with a more appropriate prior, which can facilitate the optimization, and the problem then becomes
\begin{equation}
\begin{split}
    M^*, t^* &= \mathop{\arg\max}\limits_{M,t} \log\big(p_\omega(\bar{y}|M,t)p(M)\big),\\
    &\operatorname{ s.t. }\;\ell_0(M)\leq N_0.
\end{split}
\label{eq:map}
\end{equation}

Below, we introduce an object contour as a more reasonable prior for $M$, which carries object semantics and improves the attack effectiveness, and explain the implementation details of contour acquisition, texture optimization and the MAP estimation based on the contour to further improves the attack performance.

\subsection{Using Object Contour?}

\begin{figure}[!t]
    \centering
    \includegraphics[width=\linewidth]{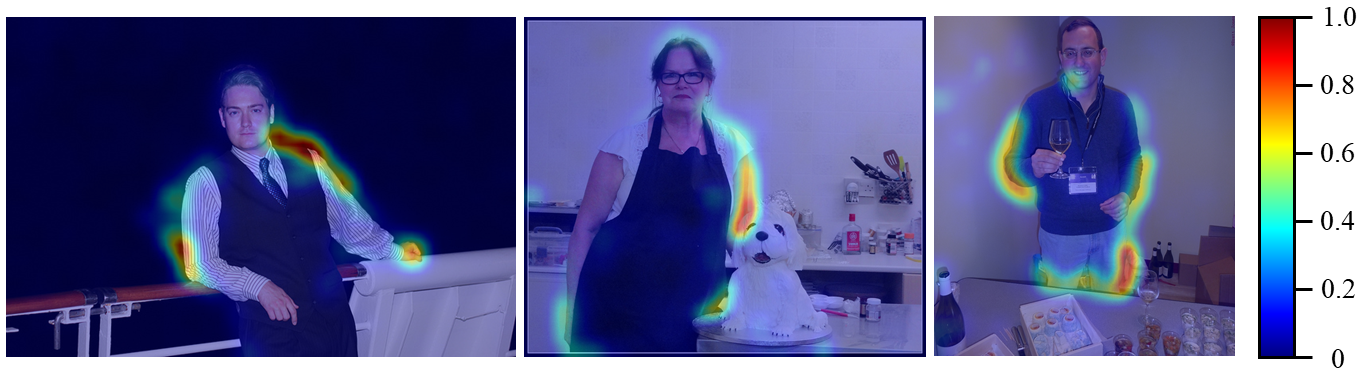}
    \caption{
    We generate heat-maps of three images on DETR by measuring the normalized adversarial contribution by replacing  different areas with adversarial noise, which indicates that the object contour is more important in adversarial attack, compared with the inside/outside regions.
    \vspace{-2ex}
    }
    \label{fig:vis}
\end{figure}
In this section, we conduct a thorough analysis and provide evidence that the object contour is not only a semantic but an informative area of the object itself, which can facilitate adversarial attacks on various object detectors. 

In~\citep{alexe2010object}, Alexe~\etal found distinctive attributes of an object and proposed different metrics closely related to color and boundary. \revise{In the analysis of shape bias and texture bias~\citep{geirhos2018imagenet}, the authors found that training a Shape-ResNet with Stylized ImageNet (SIN), which maintains shape bias while the texture is corrupted, can improve the performance in object detection. This indicates that object detectors can benefit more from shape bias when localizing the objects. Besides these studies of training better models from shape cues, we further explore the significance of contour for object detection under adversarial 
perturbations. We define a metric of normalized Adversarial Contribution ($\text{nAC}$), which measures the relative contribution of an area during adversarial attack compared with other areas in an image:}
\begin{align}
    t_a &= \mathop{\arg\max}_{t}\log p_\omega(\bar{y}|x+t\odot M_a),\\
    \text{AC}(a) &= \log p_\omega(\bar{y}|x+t_a\odot M_a)-\log p_\omega(\bar{y}|x),\\
    \text{nAC}(a) &= \frac{\text{AC}(a)-\min\limits_a \text{AC}}{\max\limits_a \text{AC}-\min\limits_a \text{AC}},
\end{align}
where $a$ represents an area on the image, and $M_a$ denotes the binary mask for area $a$. Then, we add adversarial perturbations $t_a$ to the area and evaluates $\text{AC}(a)$ by getting the difference of the model predictions caused by the added adversarial noise in $a$. And $\text{nAC}(a)$ indicates the normalized contribution among the areas across the whole image. We evaluate different contribution of an area $a$ at different area of ``Inside'', ``Contour''  or ``Outside'' corresponding to an object.

In~\cref{fig:vis}, we show the representative heat-maps describing the normalized adversarial contribution on DETR. This demonstrates that perturbations on color and texture inside the object may not be sufficient to pose any threat to object detectors with the $\ell_0$ constraint. Moreover, after examining 500 images on 9 detectors, we compare the normalized adversarial contribution of three classes of area quantitatively in \cref{tab:nac}. The results show that areas around the boundary indeed contribute more during adversarial attack, which implies that, by disturbing around the object boundary with same number of pixels, the characteristics of the object may be corrupted with higher probabilities compared with disturbing inside or outside the object.

\begin{table}[!h]
\centering
\caption{Normalized Adversarial Contribution ($\uparrow$, \%) of area examined.}
\begin{tabular}{c|c|c|c}
\hline
    & Outside&  Inside & Contour   \\ \hline
SSD    
&   7.64  & 23.58    &   \bf{25.75}  \\ 
YOLOv3          
 &   8.67  &   30.72  &  \bf{31.05} \\ 
YOLOX 
&   9.16  & 18.48    &    \bf{21.72}  \\ 
Faster R-CNN    
 &   11.49  & 23.46    &   \bf{26.77} \\
Mask R-CNN    
 &   11.61  & 22.76    &   \bf{27.00}  \\ 
Mask R-CNN Swin-T   
 &   10.06  & 17.72    &  \bf{22.68} \\ 
DETR 
&   9.38  &  11.95   & \bf{14.85}    \\ 
DAB-DETR 
&   11.65  &  15.98   &  \bf{19.03}  \\
Deformable DETR 
 &   9.33  & 19.64    &  \bf{23.95}  \\ 
 \hline
\end{tabular}
\label{tab:nac}
\end{table}

\subsection{Contour Acquisition}
\label{sec:implementation}

Here, we introduce some implementation details of our method on how to acquire the appropriate prior contour. The object semantic contour is a intuitive concept, which describes the outline of an object, which is composed of the boundary between the object and its background, and to be more precise, those between different parts of the object (\eg, those between legs, arms, head and body for human targets). Therefore, the contour acquisition can be done by intuitive approaches, like human annotations, when we are attacking a few images. As for batches of images, we can obtain the object contour from the result of instance segmentation~\citep{he2017mask}, or part segmentation~\citep{lin2020cross} if we need the contour to be more detailed. Practically, by eroding the segmentation mask, we can easily acquire sparse but highly-connected contour.

\subsection{Problem Formulation}
\label{sec:formulation}

In this section, we present our problem formulation of texture optimization based on the contour and the MAP estimation to further improve the attack performance.

\textbf{Texture Optimization.} Based on the acquired prior contour, represented as $M$, we keep it fixed and adopt a method similar to PGD~\citep{madry2017towards} to optimize the texture, which projects the updated texture $t$ into the acceptable range for pixel values as 

\begin{equation}
    t \leftarrow \textrm{Clip}_{[0,1]}\big(t + \alpha\cdot\nabla_{t} \log p_\omega(\bar{y}|M,t)\big),
\label{eq:clipattack}
\end{equation}
where we only optimize $t$ with gradients of objective function for a given $M$ and $\alpha$ is a hyper-parameter as step size. For an image with pixel values ranging from 0 to 1, we clip the modified pixel values within the acceptable range after every updating step.

\textbf{Further Optimization for Pixel Selection.} The fixed mask $M$ is likely to limit the attack performance and the further optimization can be achieved by adjusting the contour with better pixel selection. With an appropriate prior for $M$, using MAP estimation is a natural deduction from the Bayesian framework of $\ell_0$-constrained sparse attack. In particular, we may parameterize the connected contour with $\theta$, which corresponds to its binary mask representation $M(\theta)$, and obeys a prior normal distribution $\pi(\theta)$. The problem in Eq.~\eqref{eq:map} can be further reformulated as a problem parameterized by $\theta$. 
Then the optimal solution to the adversarial noise of $(\theta^\ast, t^\ast)$ can be solved via MAP estimation as  
\begin{equation}
\begin{split}
    \theta^\ast,t^\ast&=\mathop{\arg\max}\limits_{\theta,t} \log p_\omega(\bar{y}|\theta,t)+\log\pi(\theta),\\
    &\operatorname{ s.t. }\;\ell_0(M(\theta))\leq N_0,
\end{split}
\label{eq:mappp}
\end{equation}
where $\theta\sim\mathcal{N}(\theta_0,\Sigma)$ with $\theta_0$ and $\Sigma$ respectively being the prior contour and an identity matrix. Based on the prior contour with parameter of $\theta_0$, we update $\theta$ iteratively using the gradient w.r.t. $\theta$ of the expression in~\cref{eq:mappp} with $\theta_0$ as an initialization as
\begin{equation}
    \theta \leftarrow (1-\beta)\theta + \beta\big(\nabla_{\theta}\log p_\omega(\bar{y}|\theta,t)+\theta_0\big),
\end{equation}
where $\beta$ is a hyper-parameter that can be controlled to avoid $l_0(M(\theta))$ from exceeding the $\ell_0$ constraint and $t$ is the optimized texture given $\theta$. 
\begin{figure}[!t]
    \centering
    \includegraphics[width=0.95\linewidth]{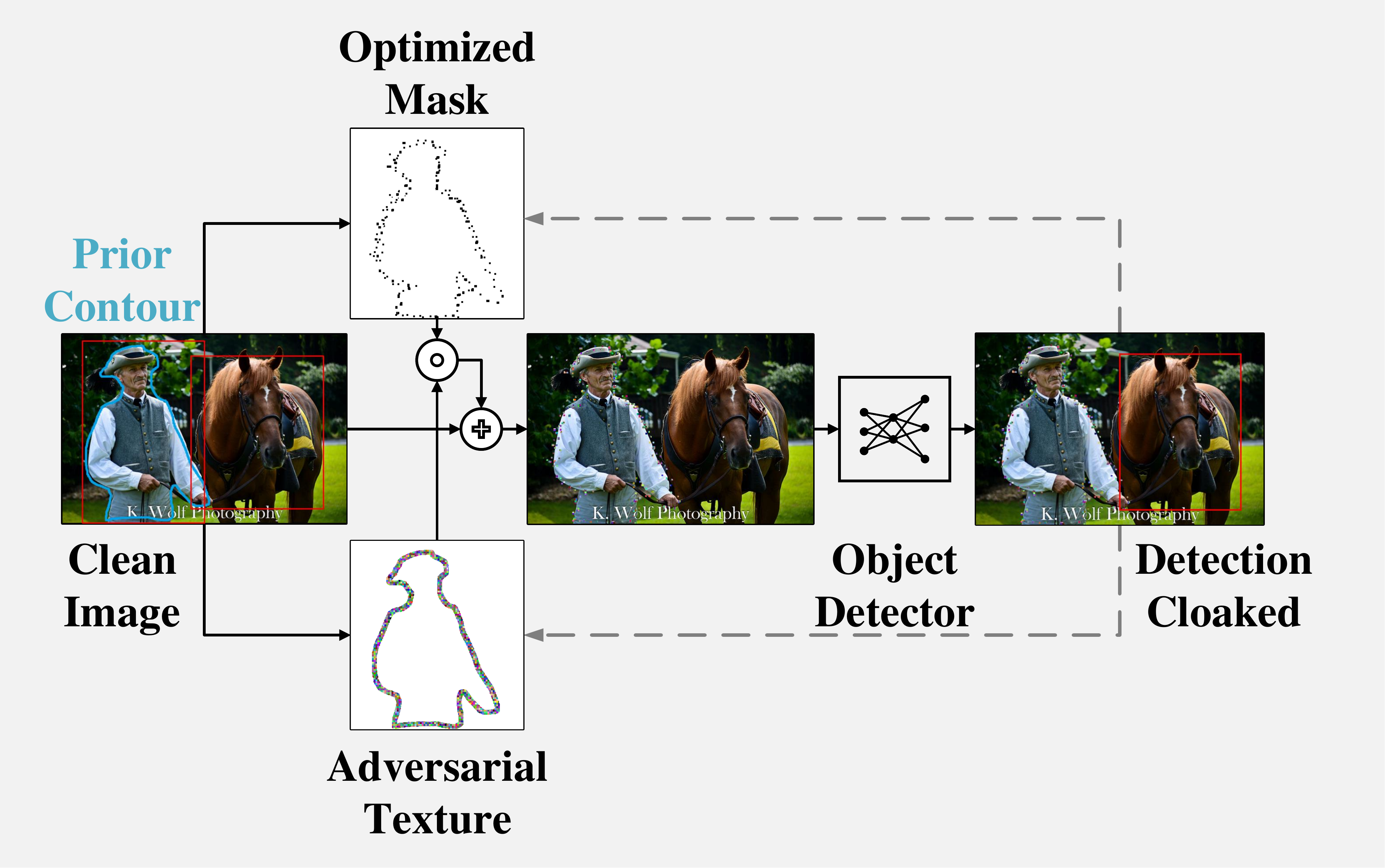}
    \caption{We separate the perturbation $\delta$ into texture $t$ and binary mask $M$. By introducing a prior distribution of $M$ around the contour, we can easily achieve a successful attack on object detectors with a MAP estimation under the Bayesian framework.
    }
    \label{fig:pipeline}
\end{figure}

However, the optimization is still intractable because the process involves the projection from continuous $\theta$ to sparse $M$. In practice, to achieve better attack successful rate, we take Monte Carlo sampling around the prior contour, which results in scattered pixels generated around the prior contour. We summarize our framework in Algorithm~\ref{alg1} and in \cref{fig:pipeline} graphically.

\begin{algorithm}[t]
\caption{Adversarial Semantic Contour} 
\label{alg1}
\begin{algorithmic}[1]
\REQUIRE inference model with weights of $\omega$; raw image $x$; label $y$; hyper-parameters $\alpha,\beta$.
\ENSURE adversarial example $x'$
\STATE Get the prior contour $C_p$ parameterized by $\theta_0$ 
\STATE Initial $\theta = \theta_0$ and $t$
\WHILE{attack fails}
\STATE Update $\theta$ as\\$\theta \leftarrow (1-\beta)\theta + \beta\big(\nabla_{\theta}\log p_\omega(\bar{y}|\theta,t)+\theta_0\big)$
\STATE Optimize $t$ for $\theta$ iteratively as \\$t \leftarrow \textrm{Clip}_{[0,1]}\big(t + \alpha\cdot\nabla_{t} \log p_\omega(\bar{y}|\theta,t)\big)$
\ENDWHILE
\STATE Project the optimized $\theta$ to binary mask $M$
\RETURN $x' \xleftarrow[]{} x+t\odot M$
\end{algorithmic}
\end{algorithm}


\begin{figure}[]
\centering

\hspace{0in}
\begin{minipage}[]{.312\linewidth}
{               
\begin{center}
    \includegraphics[width=0.6\linewidth]{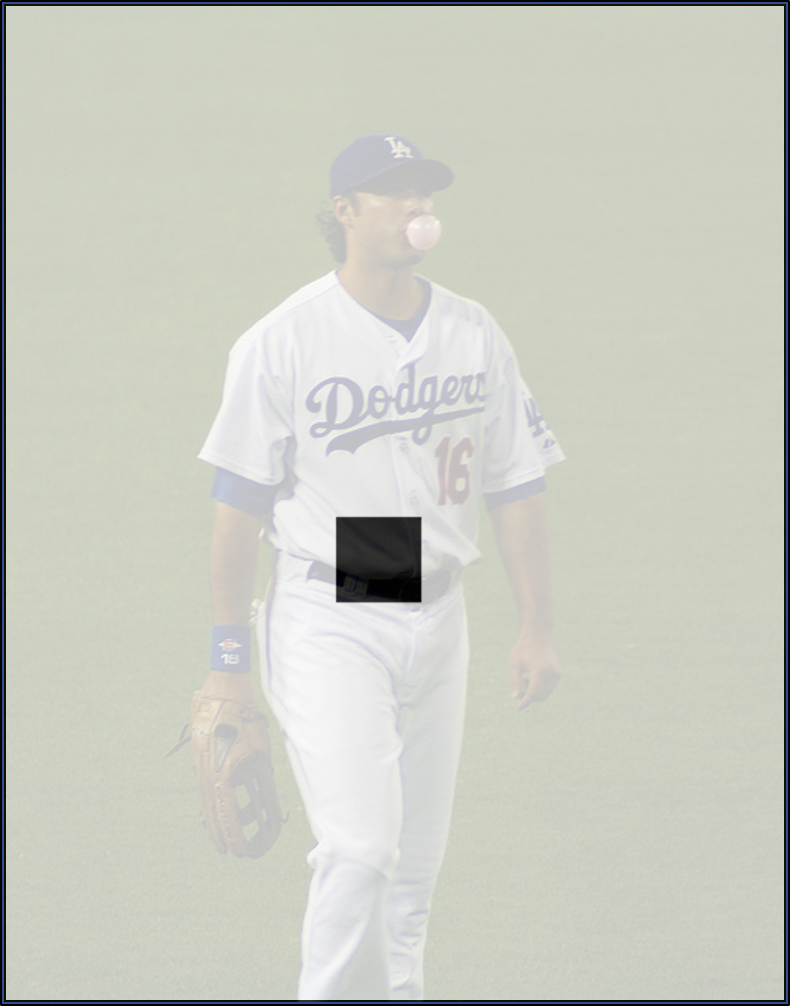}
\end{center}}
\vspace{-2ex}
\subcaption{AdvPatch}
\end{minipage}
\hspace{0in}
\begin{minipage}[]{.312\linewidth}
{               
\begin{center}
    \includegraphics[width=0.6\linewidth]{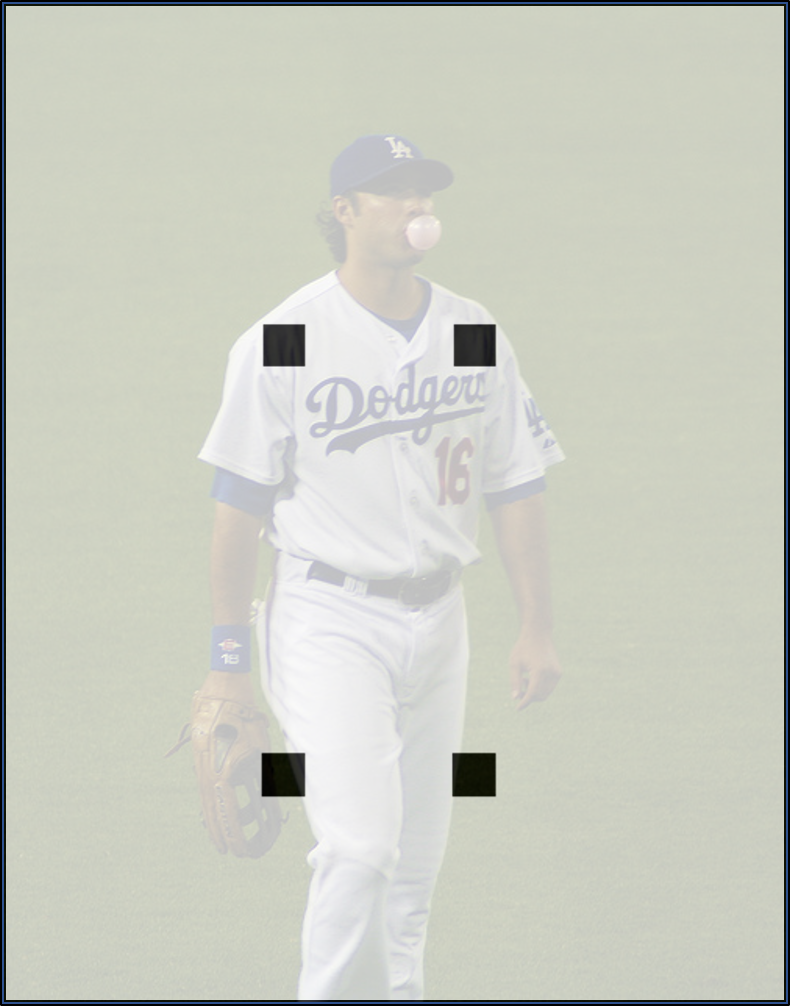}
\end{center}}
\vspace{-2ex}
\subcaption{FourPatch}
\end{minipage}
\hspace{0in}
\begin{minipage}[]{.312\linewidth}
{               
\begin{center}
    \includegraphics[width=0.6\linewidth]{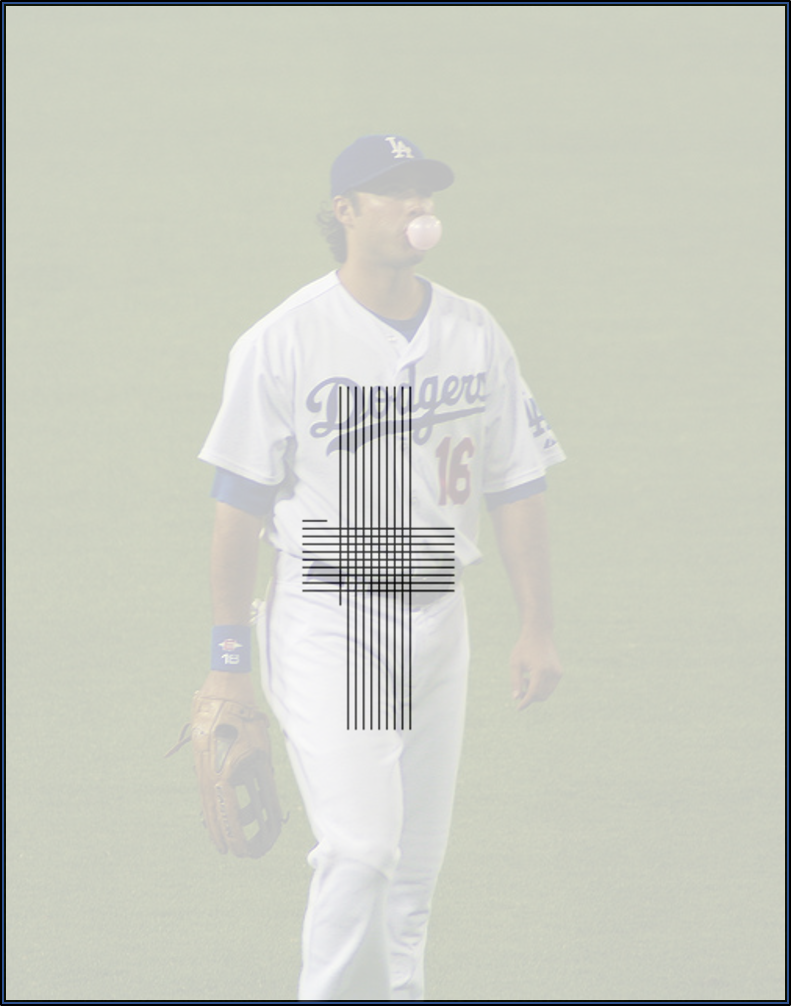}
\end{center}}
\vspace{-2ex}
\subcaption{SmallGrid}
\end{minipage}
\begin{minipage}[]{.312\linewidth}
{               
\begin{center}
    \includegraphics[width=0.6\linewidth]{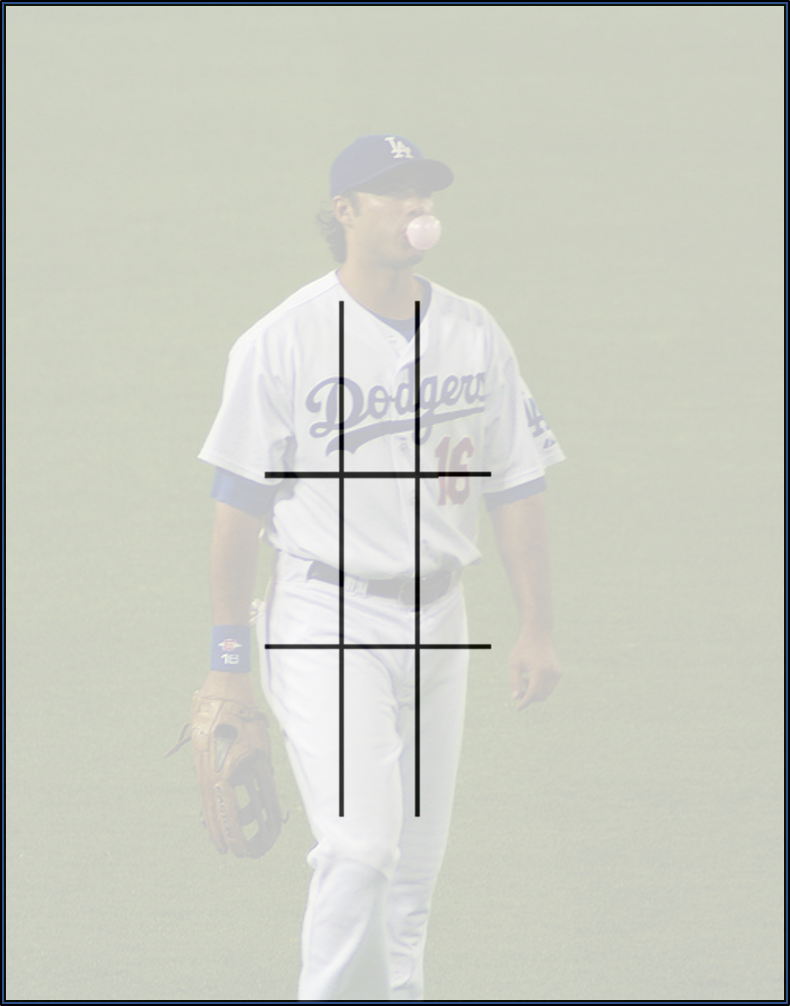}
\end{center}}
\vspace{-2ex}
\subcaption{2$\times$2Grid}
\end{minipage}
\hspace{0in}
\begin{minipage}[]{.312\linewidth}
{               
\begin{center}
  \includegraphics[width=0.6\linewidth]{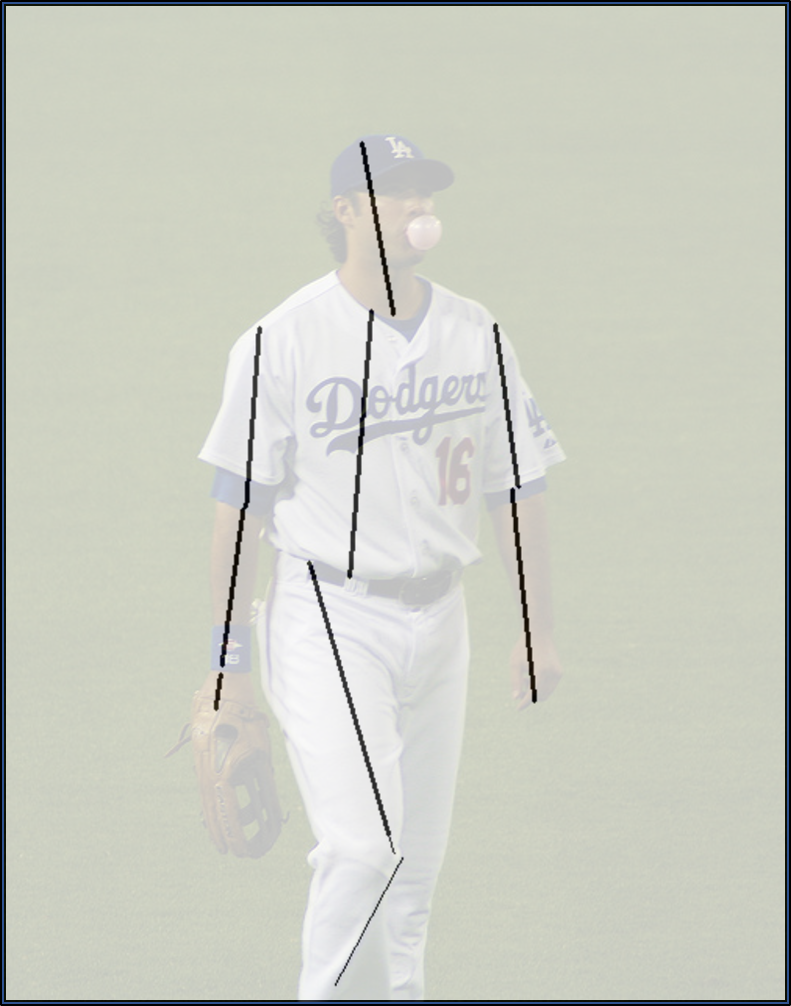} 
\end{center}}
\vspace{-2ex}
\subcaption{Strip}
\end{minipage}
\hspace{0in}
\begin{minipage}[]{.312\linewidth}
{               
\begin{center}
    \includegraphics[width=0.6\linewidth]{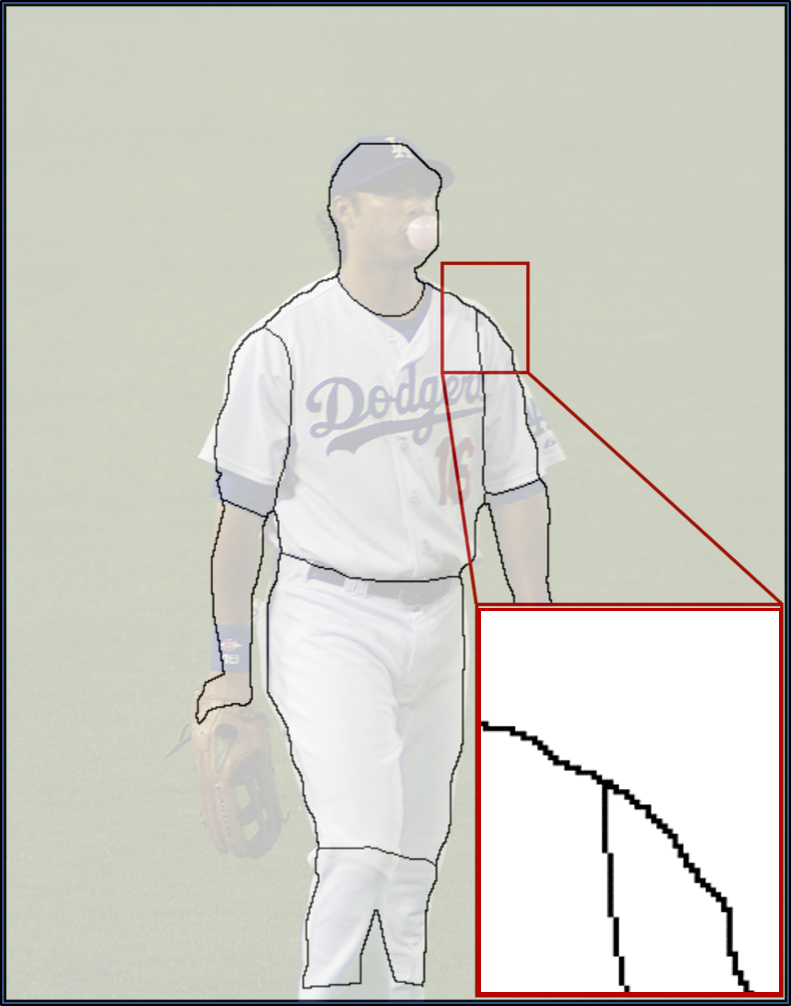}
\end{center}}
\vspace{-2ex}
\subcaption{F-ASC}
\end{minipage}
\begin{minipage}[]{.312\linewidth}
{               
\begin{center}
    \includegraphics[width=0.6\linewidth]{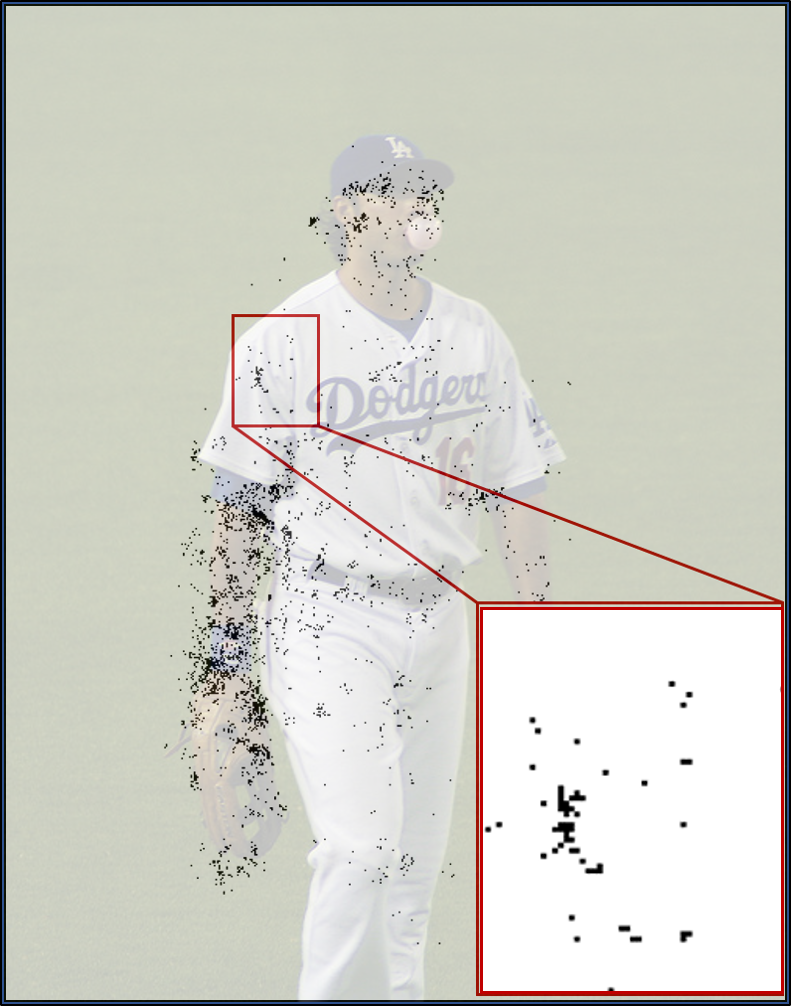}
\end{center}}
\vspace{-2ex}
\subcaption{C\&W-$\ell_0$}
\end{minipage}
\hspace{0in}
\begin{minipage}[]{.312\linewidth}
{               
\begin{center}
    \includegraphics[width=0.6\linewidth]{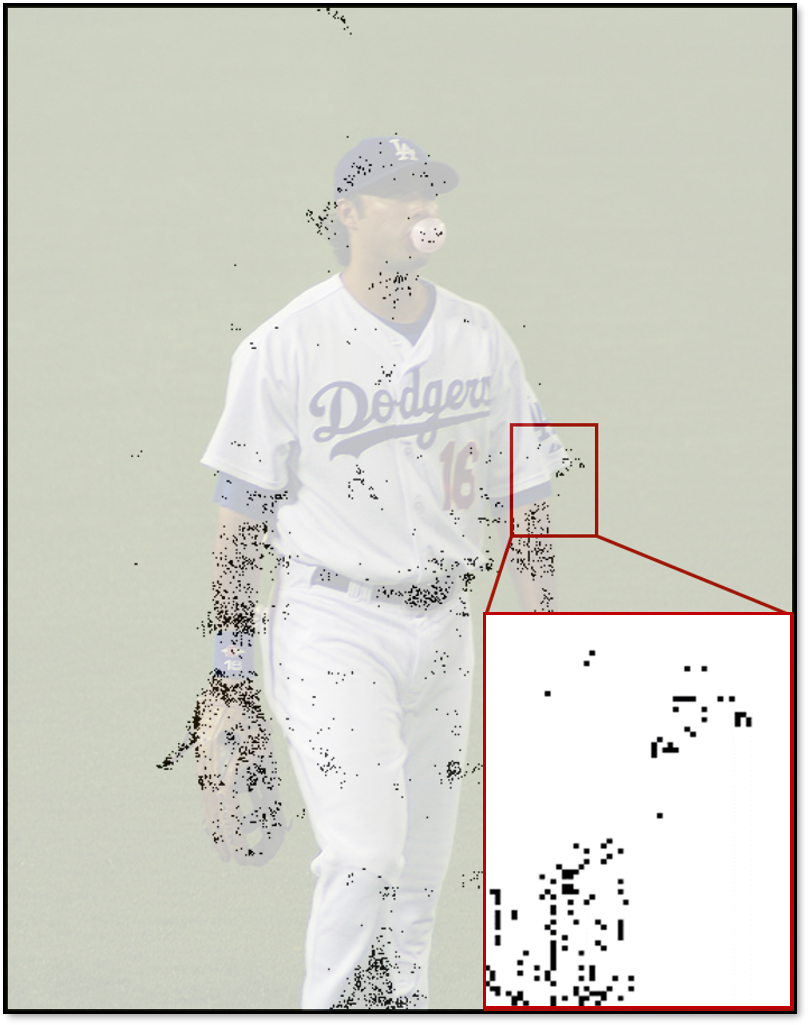}
\end{center}}
\vspace{-2ex}
\subcaption{PGD$_0$}
\end{minipage}
\begin{minipage}[]{.312\linewidth}
{               
\begin{center}
  \includegraphics[width=0.6\linewidth]{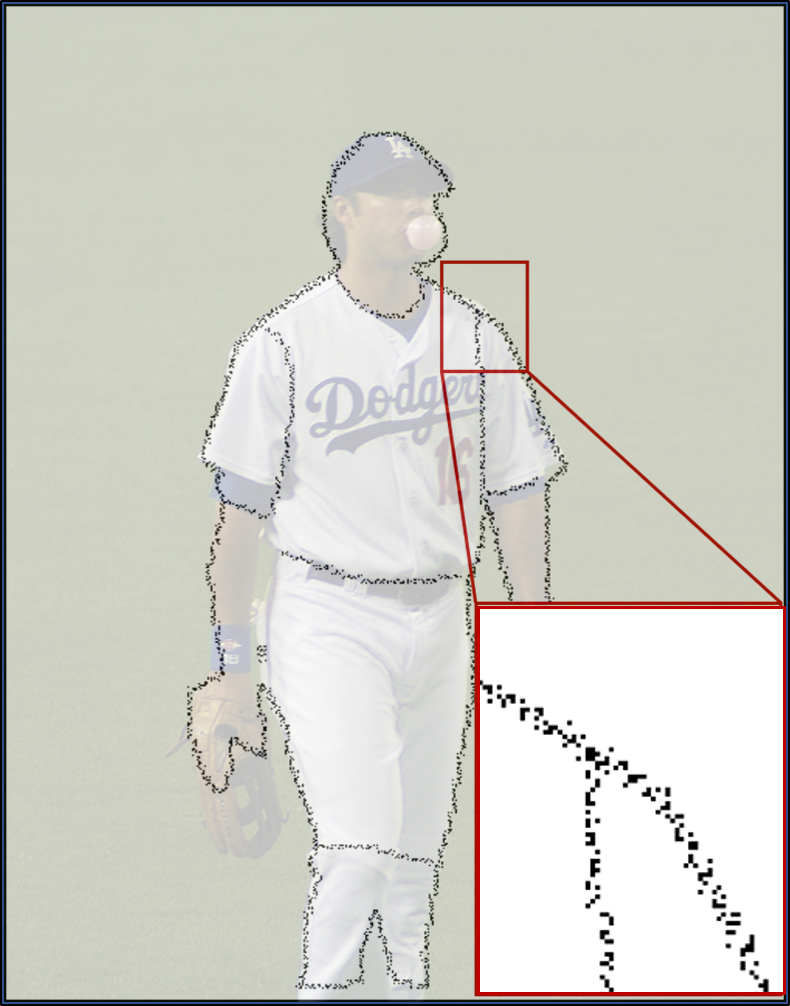}  
\end{center}}
\vspace{-2ex}
\subcaption{O-ASC}
\end{minipage}
\caption{Examples of mask patterns for  different methods.}
\label{fig:example}
\end{figure}

\section{Experiments}
\subsection{Experiment Settings}

\textbf{Datasets.} We use three datasets, including COCO~\citep{lin2014microsoft} as a universal dataset for object detection and Cityscapes~\citep{Cordts2016Cityscapes} along with BDD100K~\citep{bdd100k} as two typical datasets for autonomous driving systems.

\textbf{Models.} We mainly attack 9 models in total: 3 one-stage detectors including SSD512~\citep{Liu_2016ssd}, Yolov3~\citep{yolov3} and YoloX~\citep{yolox2021}, 3 two-stage detectors including Faster R-CNN (FRCN)~\citep{ren2015faster}, Mask R-CNN (MRCN)~\citep{he2017mask}, both with backbone of ResNet50~\citep{he2016deep} and Mask R-CNN with Swin Transformer (Swin-T)~\citep{liu2021swin} as backbone (MRCN-Swin), and 3 Detection Transformers including DETR~\citep{carion2020detr}, Deformable DETR (Def-DETR)~\citep{zhu2020deformable} and DAB-DETR~\citep{liu2022dabdetr}. These models are all trained on COCO by mmdetection~\citep{mmdetection}, except for DAB-DETR\footnote{https://github.com/IDEA-opensource/DAB-DETR}, whose official code is newly released.

\textbf{Methods.} Our method (ASC) is based on the semantic boundaries of objects. ``F-ASC'' stands for the fixed case that directly uses prior contours, which are acquired from part-segmentation~\citep{lin2020cross} for ``Person'' in COCO and ground-truth segmentation for other objects. 
Because we stress the $\ell_0$ constraints, we have the area cost of our contour pattern bounded by 5\% of the object area in COCO and 10\% in the other two datasets. We show the example patterns in~\cref{fig:example} (f) and (i). 
To verify the performance of our contour-based attack, we adopt five prior patterns as baseline methods to compare in our experiments. 
The five prior patterns are mainly in three classes, which are square patches (``AdvPatch''~\citep{thys2019fooling} and ``FourPatch'' in~\cref{fig:example} (a) and (b)), grids (``2$\times$2Grid''~\citep{wu2020dpattack} and ``SmallGrid'' in~\cref{fig:example} (c), (d)) and ``Strip'' in~\cref{fig:example} (e) generated based on part segmentation. 
We restrict the area cost of these methods to be the same as our contour pattern. \revise{Note that all methods including our proposed ASC optimize the texture via gradient-based method, which involves iterative forward inference and backward propagation, the contour acquisition for ASC with segmentation model only needs one forward inference and brings negligible extra computation under our unified attack framework.}

\begin{figure}[!t]
\centering
\begin{minipage}[]{.320\linewidth}
\footnotesize \centering
\includegraphics[width=0.98\linewidth]{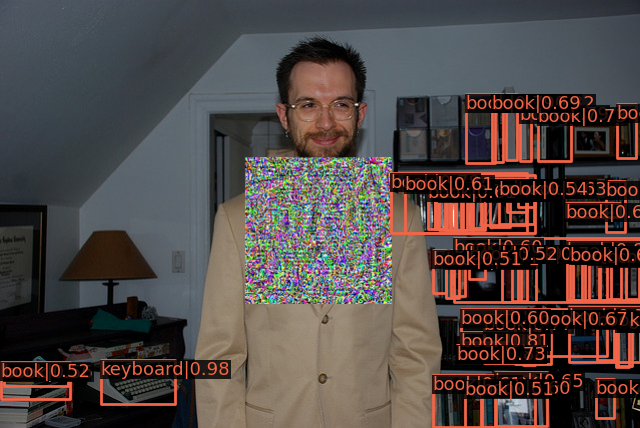}
\subcaption{AdvPatch 32.67\%}
\end{minipage}
\hspace{0in}
\begin{minipage}[]{.320\linewidth}
\footnotesize \centering
\includegraphics[width=0.98\linewidth]{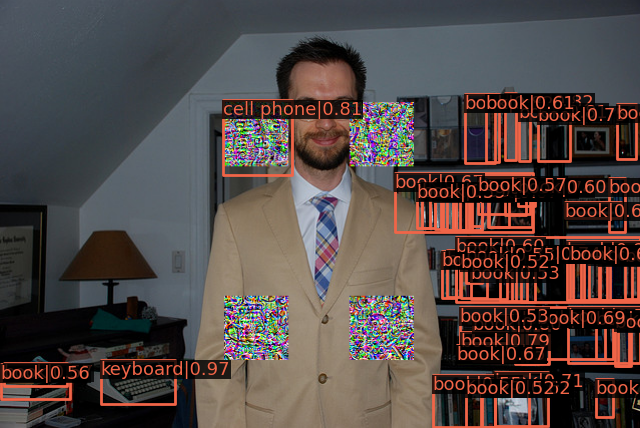}
\subcaption{FourPatch 25.11\%}
\end{minipage}
\hspace{0in}
\begin{minipage}[]{.320\linewidth}
\footnotesize \centering
\includegraphics[width=0.98\linewidth]{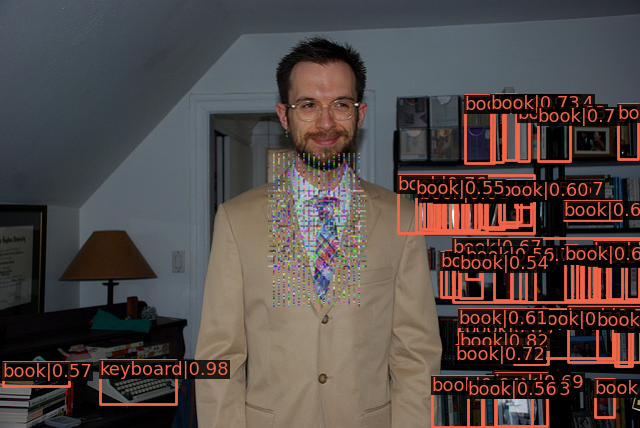}
\subcaption{SmallGrid 6.18\%}
\end{minipage}
\begin{minipage}[]{.320\linewidth}
\footnotesize \centering
\includegraphics[width=.98\linewidth]{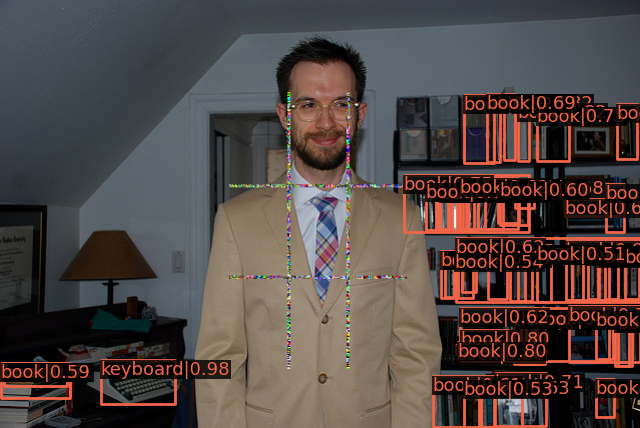}
\subcaption{2$\times$2Grid 4.11\%}
\end{minipage}
\hspace{0in}
\begin{minipage}[]{.320\linewidth}
\footnotesize \centering
\includegraphics[width=0.98\linewidth]{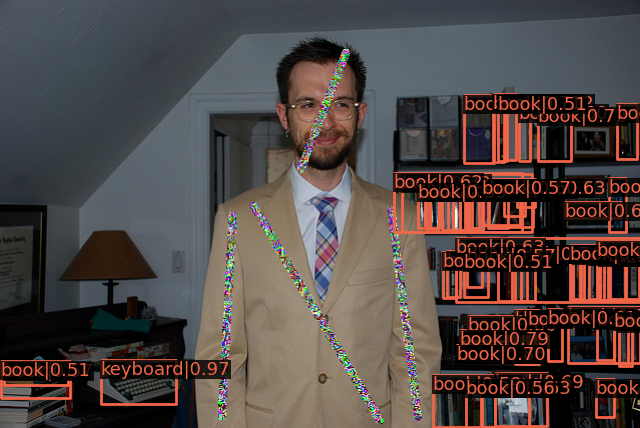}
\subcaption{Strip 9.88\%}
\end{minipage}
\hspace{0in}
\begin{minipage}[]{.320\linewidth}
\footnotesize \centering
\includegraphics[width=0.98\linewidth]{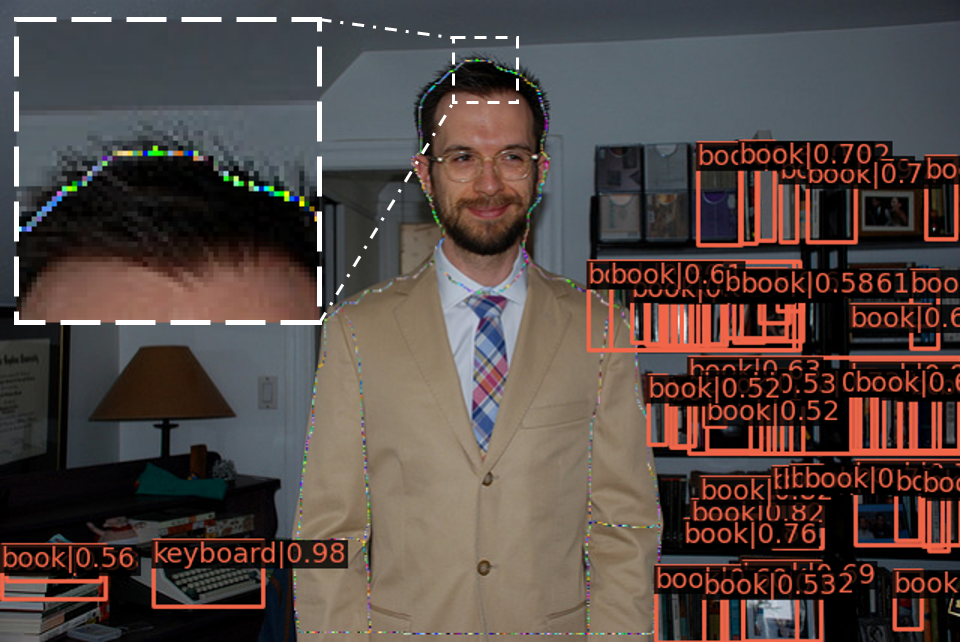}
\subcaption{F-ASC 2.94\%}
\end{minipage}
\begin{minipage}[]{.320\linewidth}
\footnotesize \centering
\includegraphics[width=\linewidth]{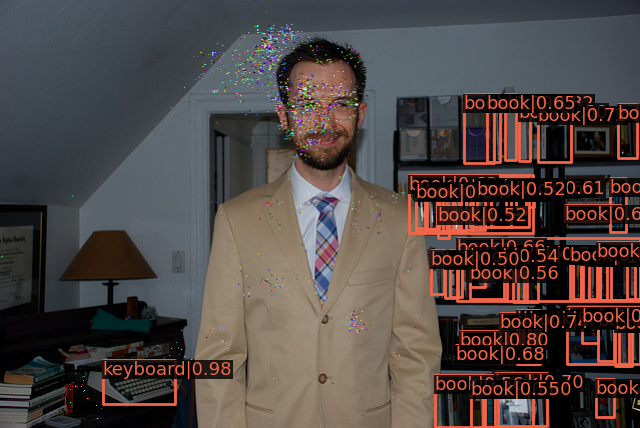}
\subcaption{C\&W-$\ell_0$ 3.50\%}
\end{minipage}
\hspace{0in}
\begin{minipage}[]{.320\linewidth}
\footnotesize \centering
\includegraphics[width=\linewidth]{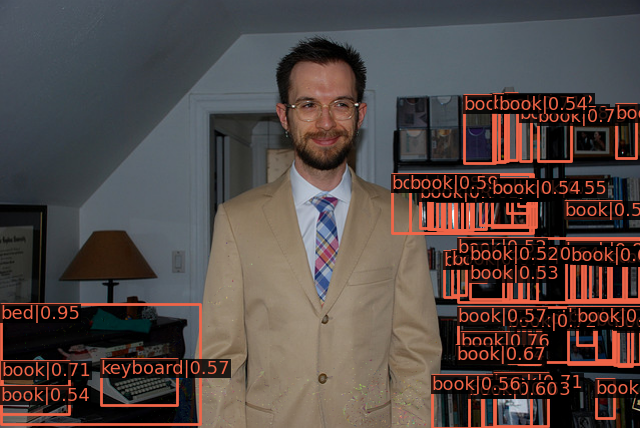}
\subcaption{PGD$_0$ 4.43\%}
\end{minipage}
\hspace{0in}
\begin{minipage}[]{.320\linewidth}
\footnotesize \centering
\includegraphics[width=0.98\linewidth]{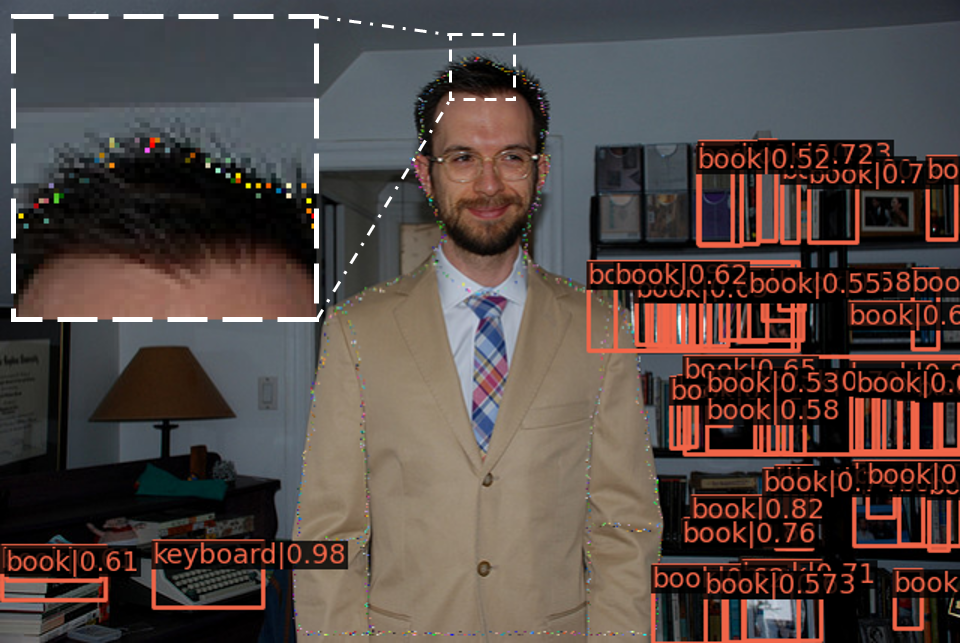}
\subcaption{O-ASC 1.78\%}
\end{minipage}
\caption{Examples of object vanishing on COCO. Our method manipulate the fewest pixels to make a person invisible to DETR. The percentage indicates the ratio of modified pixel number to pixels of the object area.}
\label{fig:cloaking}
\end{figure}

\subsection{Attack Effectiveness}

In this section, we focus on the task of object vanishing, \ie, to make the target object undetected, and verify the effectiveness of our method in terms of Successful Detection Rate (SDR), \ie, if a prediction with an IoU above 0.5 with the target object and a confidence score above a threshold (0.3 for Deformable DETR and DAB-DETR, 0.5 for the rest), then we take it as a success. We make the objects invisible to the detectors by maximizing the possibility of the object being the background, which is the most risky for the application of object detection. To boost the performance of our method, we take ``O-ASC'' for the optimized case that consists of the further pixel sampling for examples against which the attacks fail, as mentioned in Sec.\ref{sec:implementation}. Since it results in ``scattered pixels'' and the baseline patterns are all connected, to have the comparison fairer, we take two more baselines, C\&W~\citep{carlini2017towards} and PGD$_0$~\citep{croce2019sparse}, which are traditional $\ell_0$ attack methods proposed for image classification and generates perturbations over the whole image. The proposed $\ell_0$ attack in C\&W, whose binary mask is shown in~\cref{fig:example} (g), follows the pipeline that one pixel is removed in every iteration and we increase the step size for pixels for efficiency. As for PGD$_0$, which is in~\cref{fig:example} (h), we take the settings in its original paper and its application in object detection~\citep{li2020fool}.

\begin{table*}[!h]
\centering
\caption{Successful Detection Rate (\%, $\downarrow$) in object vanishing for `Person' in COCO. 
}
\begin{tabular}{l|ccccccccc}
\hline
                & SSD512    & Yolov3    & YoloX     & FRCN      & MRCN      & MRCN-Swin     & DETR      &  DAB-DETR     & Def-DETR          \\ \hline
\hline  
Clean           & 95.6      & 92.1      &  99.4     & 98.3      & 98.9      &   99.6        & 99.6      &   98.5        &   97.3           \\
\hline  
AdvPatch        & 6.2       & 11.5      &   25.2    & 85.0      & 88.7      &   81.7        & 94.5      &  74.7         &       67.1        \\
FourPatch       & 22.8      & 17.0      &    43.3   & 76.4      & 77.7      &   79.5        & 90.4      &  67.1         &       46.2        \\
SmallGrid       & 2.0       & 0.4       &    5.6    & 46.9      & 53.3      &   51.5        & 85.7      &    28.1       &   20.0            \\ 
2$\times$2Grid  & 0.9       &  \bf{0.2} &     5.3   & 28.3      & 32.9      &     41.0      & 75.3      &    18.5       &    7.8             \\
Strip           & 1.1       & 0.8       &   3.6     & 16.8      & 21.4      &   45.3        & 47.9      &  9.9          &    4.4            \\ 
F-ASC (ours)    & \bf{0.3}  & 0.7       &  \bf{1.7} & \bf{9.7}  & \bf{10.8} &   \bf{24.4}   & \bf{25.3} &  \bf{4.0}     &    \bf{2.6}            \\ \hline \hline
C\&W-$\ell_0$   & 0.2       & 0.6       &  2.3      & 13.1      & 15.1      &    29.4       & 46.6      &   5.4         &       1.8            \\ 
PGD$_0$         & 0.4       & 0.3       &   8.4     & 8.8       & 9.3       &   22.4        & 21.8      &   6.6         &    3.1               \\
O-ASC  (ours)   & \bf{0.1}  & \bf{0.1}  &   \bf{0.2}& \bf{2.0}  & \bf{2.7}  &   \bf{10.7}   &\bf{5.2}   &   \bf{1.0}    &   \bf{0.3}                \\ \hline
\end{tabular}
\label{tab:disappear_person}

\end{table*}

\begin{table*}[!h]
\centering
\caption{Successful Detection Rate (\%, $\downarrow$) in object vanishing for other categories in COCO. 
}
\begin{tabular}{l|ccccccccc}
\hline
                & SSD512    & Yolov3    & YoloX     & FRCN   & MRCN     & MRCN-Swin     & DETR.     & DAB-DETR      &    Def-DETR       \\ \hline
\hline
Clean           & 90.1      & 89.4      &  96.2     & 90.7   & 92.0     &   97.8        & 97.9      &  93.3         &   93.2            \\
\hline
AdvPatch        & 5.7       & 13.2      &   26.1    & 57.3   & 62.6     &   68.8        & 71.8.     &  45.6         &  47.4                 \\
FourPatch       & 12.2      & 12.4      &  30.5     & 45.3   & 48.8     &   69.5        & 63.3      &  38.0         &  26.6                 \\
SmallGrid       & 3.0       & 1.9       &  6.6      & 20.8   & 22.5     &    31.2       & 53.1.     &   14.5        &   11.8            \\ 
2$\times$2Grid  & \bf{2.8}  &  \bf{1.0} &   4.2     & 12.1   & 13.0     &    28.9       & 45.0      &   9.7         &   6.9             \\
Strip           & 6.5       & 4.0       &  13.5     & 25.7   & 26.7     &  44.8         & 47.5      &  16.2         &   9.9             \\
F-ASC (ours)    & 3.5       & 2.9       &   \bf{4.0}&\bf{9.4}& \bf{9.2} &    \bf{25.2}  & \bf{18.8} &  \bf{5.3}     &    \bf{2.0}            \\ \hline \hline
C\&W-$\ell_0$   & \bf{1.0}  & \bf{0.4}  &  1.6      & 8.7    & 8.0      &  22.1         & 29.8      &   3.0         &       1.2         \\ 
PGD$_{0}$       & 2.7       & 1.0       &    5.0    & 5.1    & 7.0      &    19.0       & 18.0      &  4.1          &     3.4           \\
O-ASC (ours)    & 1.8       & 1.0       &   \bf{0.7}&\bf{3.8}&\bf{3.5}  &  \bf{13.5}    & \bf{ 5.6} &   \bf{0.9}    &    \bf{0.7}               \\ \hline
\end{tabular}
\label{tab:disappear_other}

\end{table*}

\textbf{Experiments on COCO.} 
COCO~\citep{lin2014microsoft} is a popular dataset with 80 categories for object detection and other visual tasks. We randomly select 1000 images with labels of ``Person'' and 1000 of other categories as the attack targets.

\revise{To better visualize the margins of attack performance within the methods as in~\cref{fig:cloaking}, we examine the minimum $\ell_0$ cost to achieve successful attacks against DETR~\citep{carion2020detr} for each method, by increasing $l_0$ constraint at certain step size until the person is cloaked. This has a different setting from the following quantitative experiments with same $l_0$ budget, but their conclusions of effectiveness are consistent.} 
Our methods, both ``F-ASC'' with fixed contour and ``O-ASC'' with optimized selection, require far fewer pixels to make the person invisible, indicating the attack effectiveness of our method. It is noted that compared to ``AdvPatch,'' which is the most popular pattern, our methods reduce the number of modified pixels from around 30\% to less than 3\%.

Quantitative experimental results are shown in \cref{tab:disappear_person} and \cref{tab:disappear_other}, which are for ``Person'' and other categories in COCO respectively. We list the results of these two types of objects separately because the prior contour for the ``Person'' is acquired from part segmentation, which is more precise and detailed, while that for the other is only from ground-truth segmentation label. Both tables indicate the similar trends. We see that for one-stage detectors, their adversarial robustness are the weakest and most of the methods can lead to quite low SDRs. Our methods achieve lower or similar SDRs on these detectors, both for ``F-ASC'' compared to the other five patterns and for ``O-ASC'' compared to the two traditional $\ell_0$ attack methods. As for the other two types of detectors, two-stage detectors and Detection Transformers, our methods can lead to the lowest SDRs with quite significant gaps. To be specific, the gaps of SDR between ``F-ASC'' and other prior patterns on DETR are more than 22.6\% for ``Person'' and 26.2\% for other categories, and those between ``O-ASC'' and two $\ell_0$ attack methods on it are more than 16.6\% for ``Person'' and 12.4\% for other categories. The reason is that the object contour serves as a common feature of the object among the detectors and carries more information of the object existence, which can mislead various detectors. Also, with further sampling, ``O-ASC'' is much more effective than ``F-ASC''. For instance, on DETR, the declines in SDRs are 10.1\% and 13.2\% respectively. This indicates the update of pixel selection based on the prior contour can be promising.

\begin{table*}[!h]
\centering
\caption{Successful Detection Rate (\%, $\downarrow$) in Cityscapes. 
}
\begin{tabular}{l|ccccccccc}
\hline
               & SSD512 & Yolov3 & YoloX & FRCN  & MRCN  & MRCN-Swin & DETR & DAB-DETR & Def-DETR \\ \hline
Clean          & 99.3   & 93.6   & 99.9  & 100.0 & 100.0 & 100.0     & 99.9 & 99.9     & 99.9     \\ \hline
AdvPatch       & 15.8   & 31.1   & 41.1  & 80.1  & 82.6  & 88.3      & 90.4 & 80.7     & 64.5     \\
FourPatch      & 13.8   & 24.1   & 43.1  & 57.0  & 61.6  & 85.9      & 79.0 & 47.6     & 30.3     \\
SmallGrid      & 5.2    & 8.5    & 4.1   & 39.6  & 40.8  & 60.7      & 78.1 & 20.8     & 20.6     \\
2$\times$2Grid & 3.1    & 6.4    & 4.8   & 22.9  & 31.6  & 63.8      & 71.8 & 5.8      & 6.8      \\
Strip          & 3.1    & 11.8   & 14.6  & 30.4  & 37.9  & 72.4      & 63.5 & 13.7     & 6.9      \\
F-ASC(ours)    & \textbf{2.3}    & \textbf{3.0}    & \textbf{3.5}   & \textbf{21.4}  & \textbf{30.5}  & \textbf{55.6}      & \textbf{22.4} & \textbf{1.6}      & \textbf{0.2}      \\ \hline
C\&W-$\ell_0$  & \textbf{0.1}    & 0.6    & 2.1   & 4.7   & 7.6   & 38.6      & 36.7 & 1.4      & 1.3      \\
PGD$_0$        & 5.3    & 1.1    & 0.4   & 11.5  & 13.1  & 37.4      & 31.4 & 2.6      & 1.8      \\
O-ASC(ours)    & \textbf{0.1}    & \textbf{0.2}    & \textbf{0.2}   & \textbf{2.9}   & \textbf{6.1}   & \textbf{26.3}      & \textbf{2.1}  & \textbf{0.2}      & \textbf{0.0}      \\ \hline
\end{tabular}
\label{tab:cityscapes}
\end{table*}

\begin{table*}[!h]
\centering
\caption{Successful Detection Rate (\%, $\downarrow$) in BDD100K. 
}
\begin{tabular}{l|ccccccccc}
\hline
   & SSD512 & Yolov3 & YoloX & FRCN & MRCN & MRCN-Swin & DETR  & DAB-DETR & Def-DETR \\ \hline
Clean    
& 93.4   
& 83.1   & 98.8  & 98.1 & 98.5 & 99.2& 98.3  & 97.0  & 97.5   \\ \hline
AdvPatch 
& 23.4   & 21.7   & 49.8  & 79.6 & 81.1 & 81.9& 91.1 & 77.2   & 66.3   \\
FourPatch
& 24.8   & 8.3    & 41.4  & 61.2 & 63.9 & 78.9& 77.8 & 51.7  & 34.7     \\
SmallGrid
& 4.0    & 2.2    & 9.0   & 39.6 & 42.5 & 52.8& 77.5  & 19.6  & 13.4   \\
2$\times$2Grid 
& 2.8    & 1.0    & 9.6   & 17.5 & 20.1 & 51.4& 64.6  & 5.9 & 3.9 \\
Strip    
& 6.8    & 1.9    & 17.5  & 24.1 & 26.4 & 58.4& 59.8  & 12.2  & 5.4   \\
F-ASC(ours)    
& \textbf{2.7}    & \textbf{0.8}    & \textbf{6.5}   & \textbf{14.8} & \textbf{18.9} & \textbf{51.1} & \textbf{29.6}  & \textbf{2.8} & \textbf{1.0} \\ \hline \hline
C\&W-$\ell_0$  
& \textbf{0.0}    & \textbf{0.0}    & 1.3   & 7.5  & 10.7 & 33.5& 38.6 & 1.7 & 1.4\\
PGD$_0$  
& 4.6    & 0.6    & 2.0   & 8.7  & 10.4 & 37.7& 43.7 & 4.3 & 4.3 \\
O-ASC(ours)    
& 0.7    & \textbf{0.0}    & \textbf{1.0}   & \textbf{6.5}  & \textbf{8.7}  & \textbf{31.8} & \textbf{8.0}   & \textbf{1.1} & \textbf{0.4} \\ \hline
\end{tabular}
\label{tab:bdd100k}
\end{table*}

Two noticeable phenomena appear in our results. The first one is that, with Swin-T as the backbone network, the adversarial robustness of Mask R-CNN is improved compared to that with ResNet50. The resulting SDRs of almost all methods become higher, indicating that the Swin-T backbone help Mask R-CNN resist the adversarial attack. It has been studied that ViT models have global receptive fields, which allows to model global context and therefore lead to robustness of learned features~\citep{naseer2021intriguing}. Swin Transformer, though it computes self-attention within local windows to improve efficiency, still have larger receptive fields (window size of at least $7\times7$) than ResNet50 (kernel size of $3\times3$)~\citep{liu2022convnet} and results in better robustness to our attack. The second one is that, among the three Detection Transformers, the two incremental versions of DETR appear to be much easier to be attacked, which suggests their improvement measures to uplift the prediction accuracy in terms of mean Average Precision (mAP) bring with worse adversarial robustness. Concretely, they are DETR, DAB-DETR and Deformable DETR, arranged from the most robust to the weakest. One typical difference between DETR and its incremental versions is that DETR uses Cross Entropy Loss while the other two models use Focal Loss~\citep{lin2017focal}. This lead to better predicting performance but sacrifice the robustness, maybe because Focal Loss is proposed based on Binary Cross Entropy Loss to address the imbalance of positive and negative samples, and to put more weights on objects that are harder to distinguish. Different sizes of receptive fields may also contribute to this phenomenon. DETR~\citep{carion2020detr} adopts 
cross-attention mechanism whose receptive field covers the whole feature map, while Deformable DETR~\citep{zhu2020deformable} employs deformable attention with sparse sampling points which significantly decrease the size of receptive field. As for DAB-DETR~\citep{liu2022dabdetr}, it uses the scale information of dynamic anchor boxes to modulate the attention, similar to soft RoI pooling, which leads to its receptive field smaller than DETR but larger than Deformable DETR. The rank of the equivalent receptive fields corresponds to that of their robustness to our attack.

Moreover, the separate results of ``Person'' and other categories in the two tables again confirm the correctness of our claim about object semantics. For meaningless pre-designed patterns, C\&W-$\ell_0$ and PGD$_0$, most of their resulting SDRs for ``Person'' are higher than those for other categories. This indicates that detectors tend to be more accurate and robust when detecting humans. However, this kind of gap is much smaller for ``Strip'' and ASC. Since ``Strip'' and object contour are acquired from part segmentation for ``Person'', which is more precise and thereby carries more accurate semantic information, they can still achieve similar or even better attack performance than that of other categories.

\begin{figure}[!t]
\centering
\begin{minipage}[]{.48\linewidth}
{               
\includegraphics[width=\linewidth]{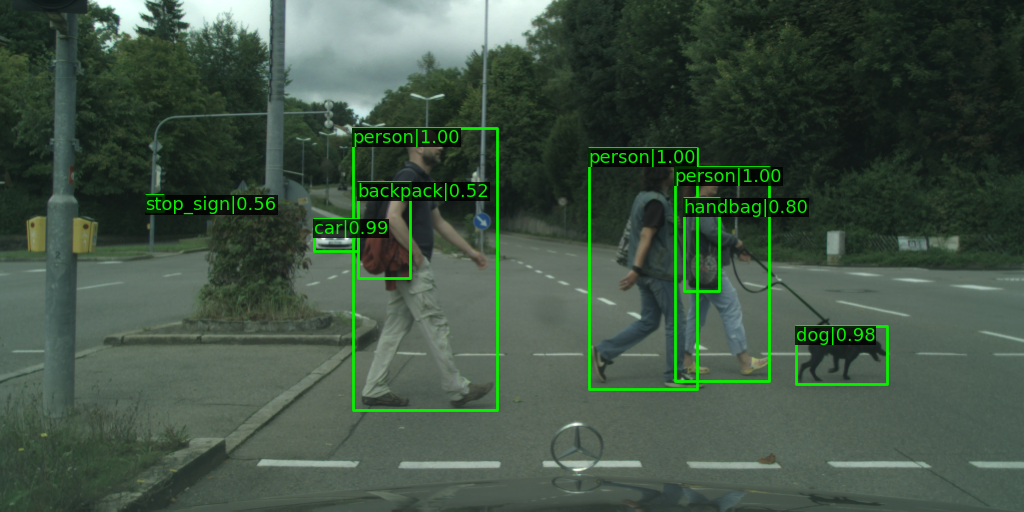}}
\subcaption{Successful Detection}
\end{minipage}
\hspace{0in}
\begin{minipage}[]{.48\linewidth}
{               
\includegraphics[width=\linewidth]{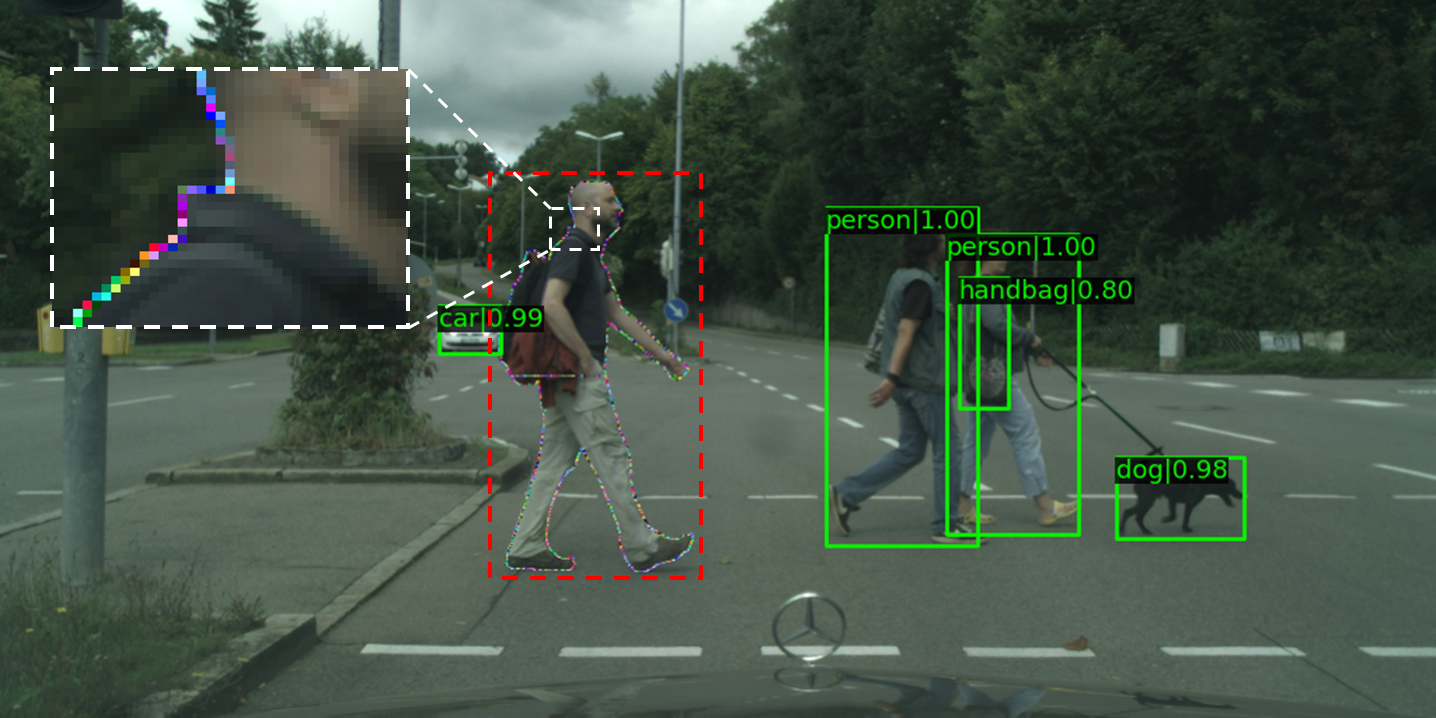}}
\subcaption{Invisible}
\end{minipage}
\begin{minipage}[]{.48\linewidth}
{               
\includegraphics[width=\linewidth]{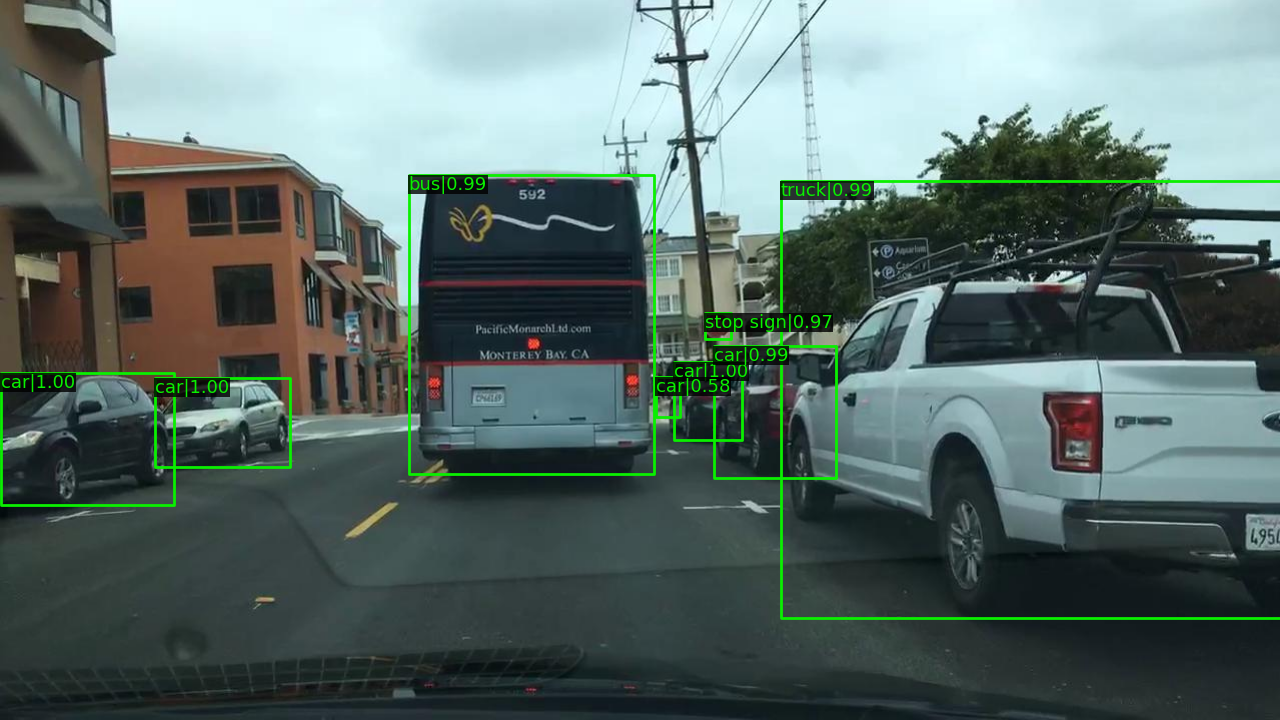}}
\subcaption{Successful Detection}
\end{minipage}
\hspace{0in}
\begin{minipage}[]{.48\linewidth}
{               
\includegraphics[width=\linewidth]{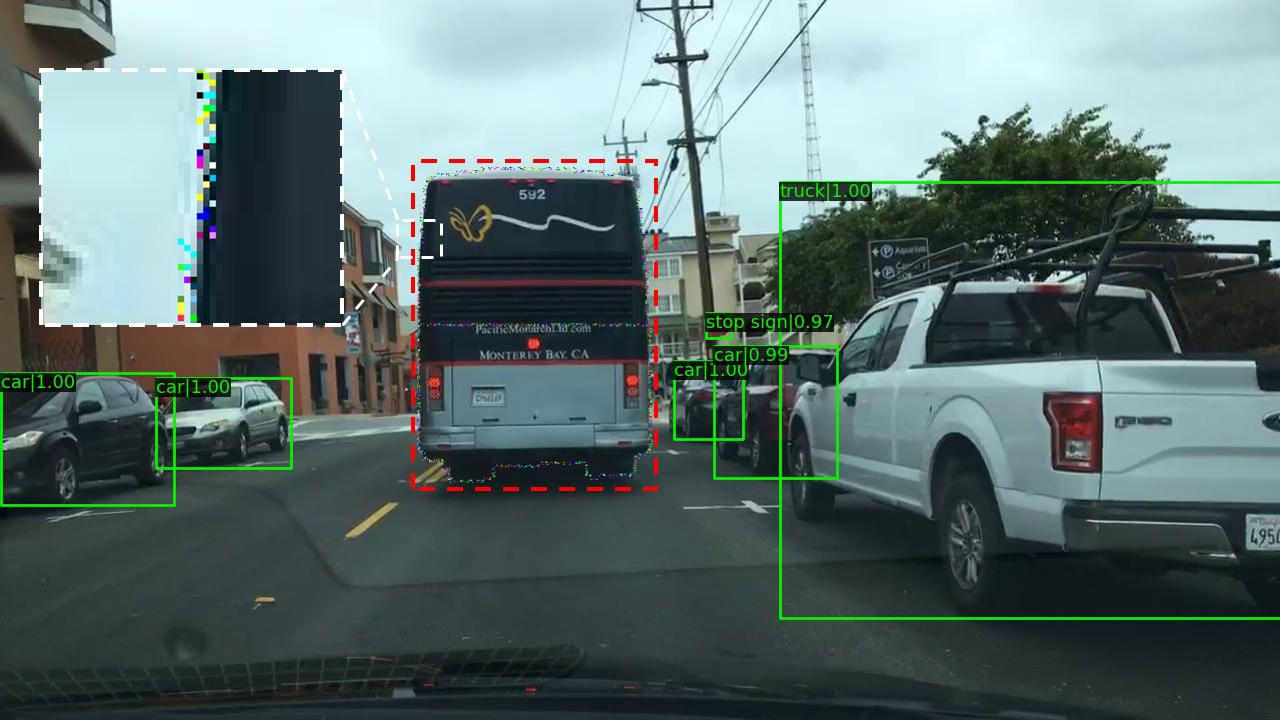}}
\subcaption{Invisible}
\end{minipage}
\caption{Examples of our attack on Cityscapes and BDD100K.
(a) and (b) are images from Cityscapes, (c) and (d) are images from BDD100K. The red boxes with dash line highlight that the objects are invisible to the detectors.
}
\label{fig:auto-driving}
\end{figure}

\begin{figure*}[!h]
\centering
\begin{minipage}[]{.19\linewidth}
{               
\includegraphics[width=\linewidth]{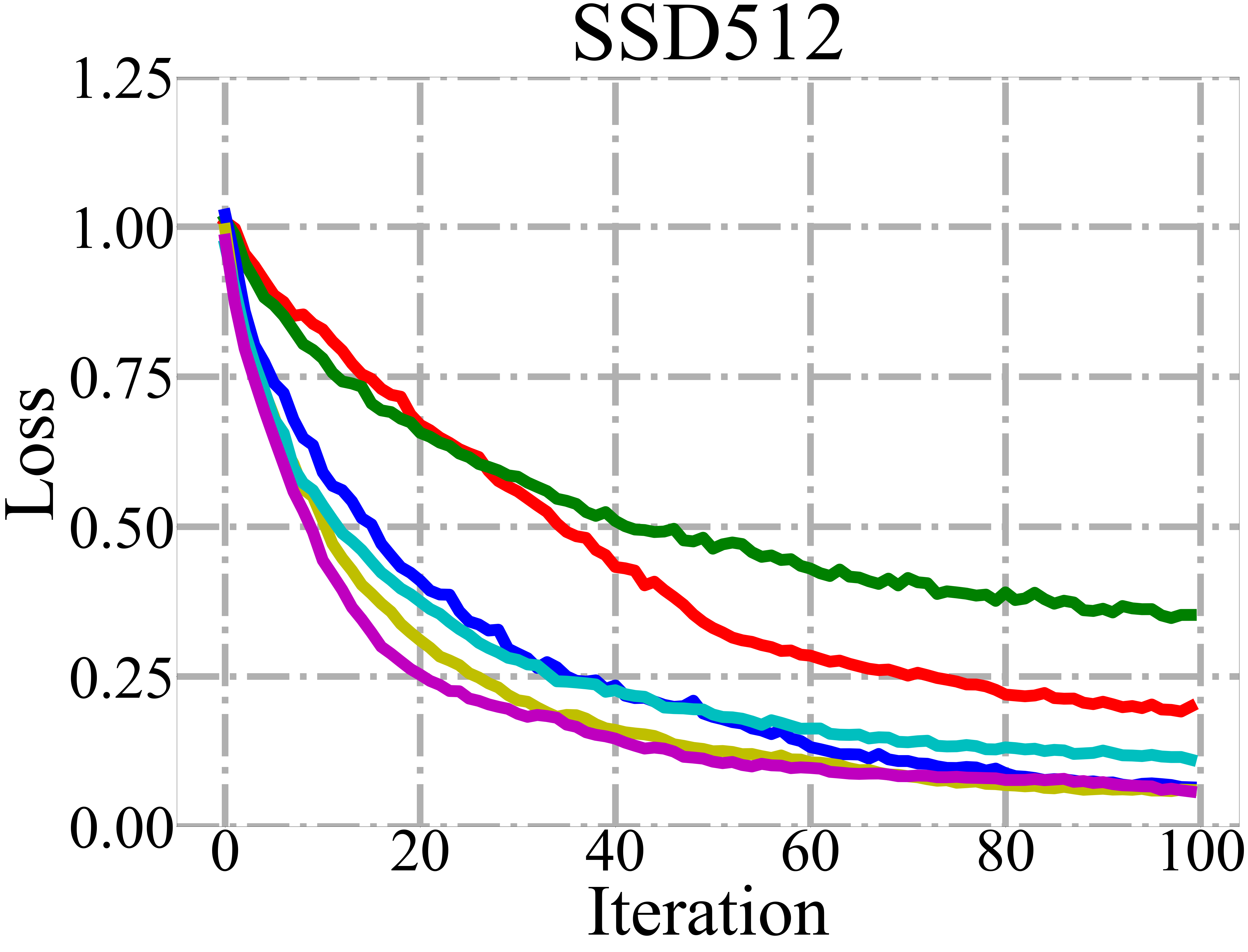}}
\end{minipage}
\hspace{0in}
\begin{minipage}[]{.19\linewidth}
{               
\includegraphics[width=\linewidth]{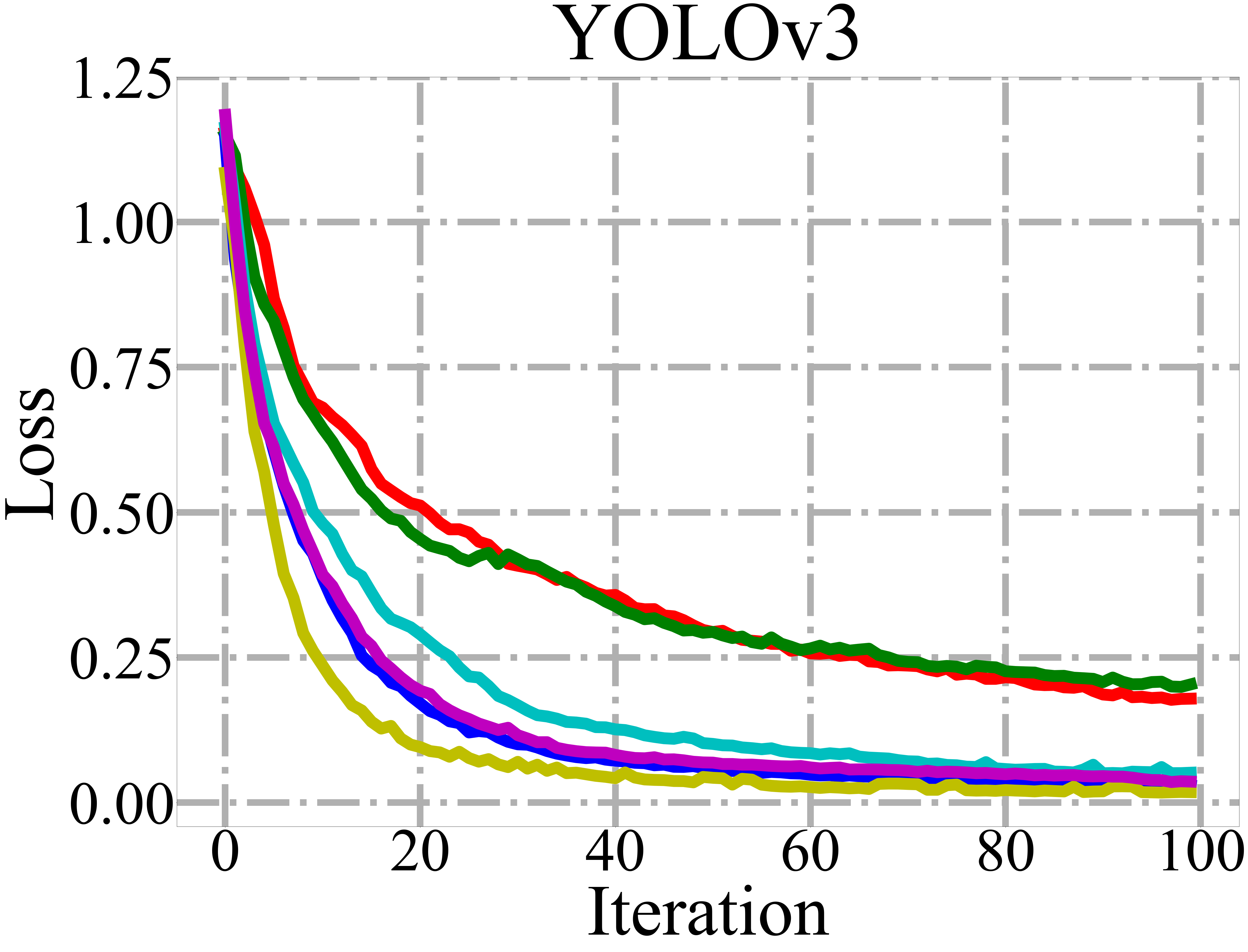}}
\end{minipage}
\hspace{0in}
\begin{minipage}[]{.19\linewidth}
{               
\includegraphics[width=\linewidth]{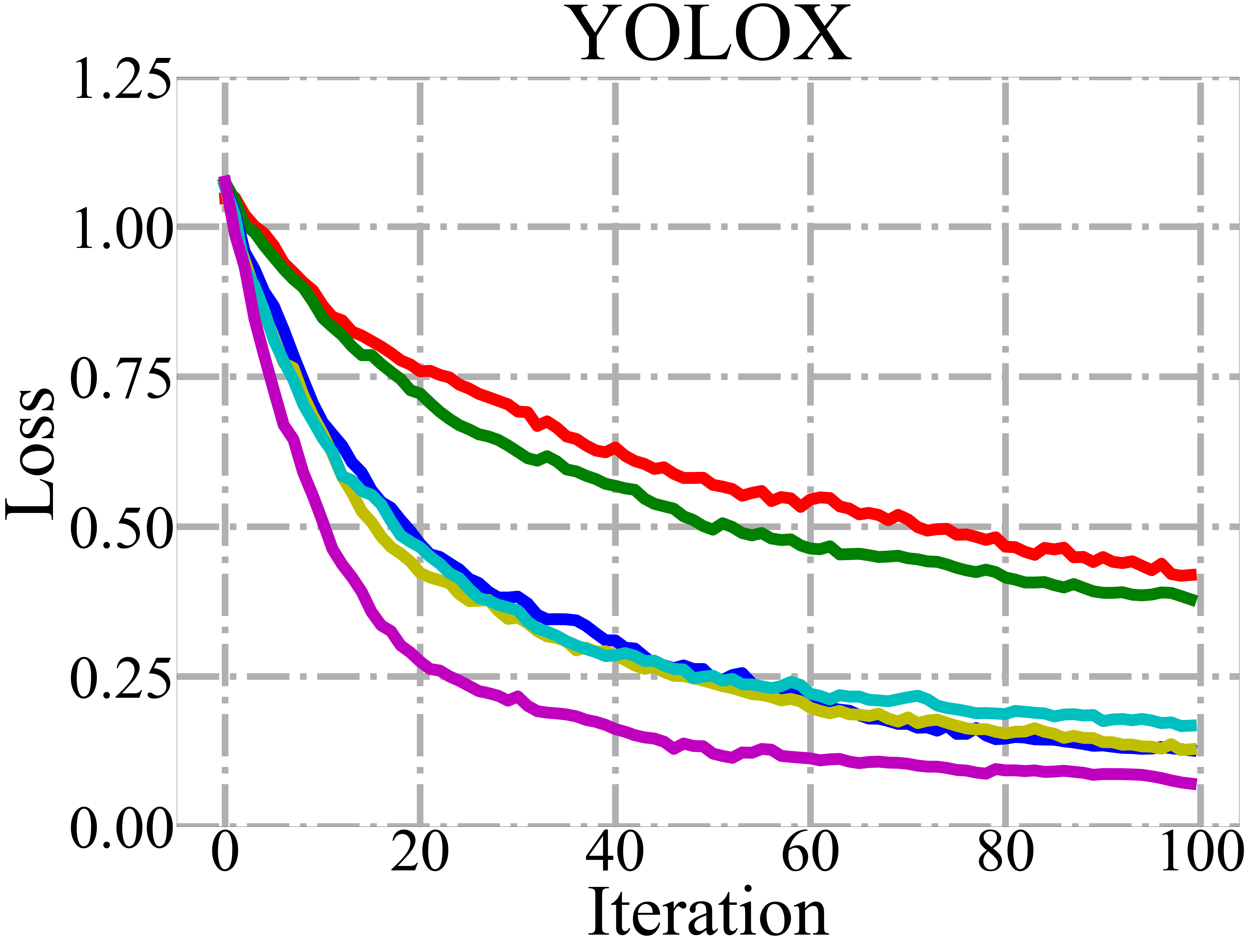}}
\end{minipage}
\hspace{0in}
\begin{minipage}[]{.19\linewidth}
{               
\includegraphics[width=\linewidth]{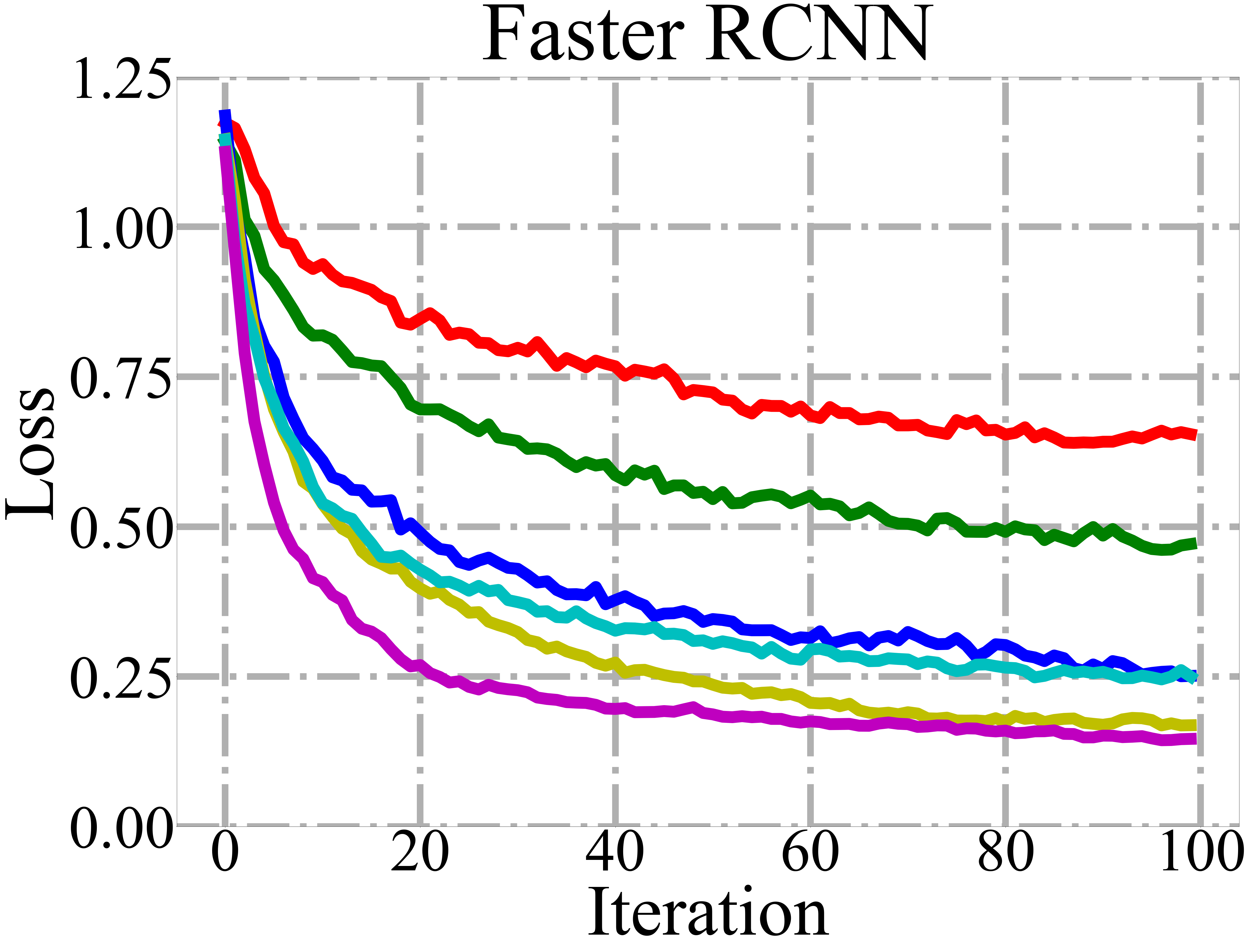}}
\end{minipage}
\hspace{0in}
\begin{minipage}[]{.19\linewidth}
{               
\includegraphics[width=\linewidth]{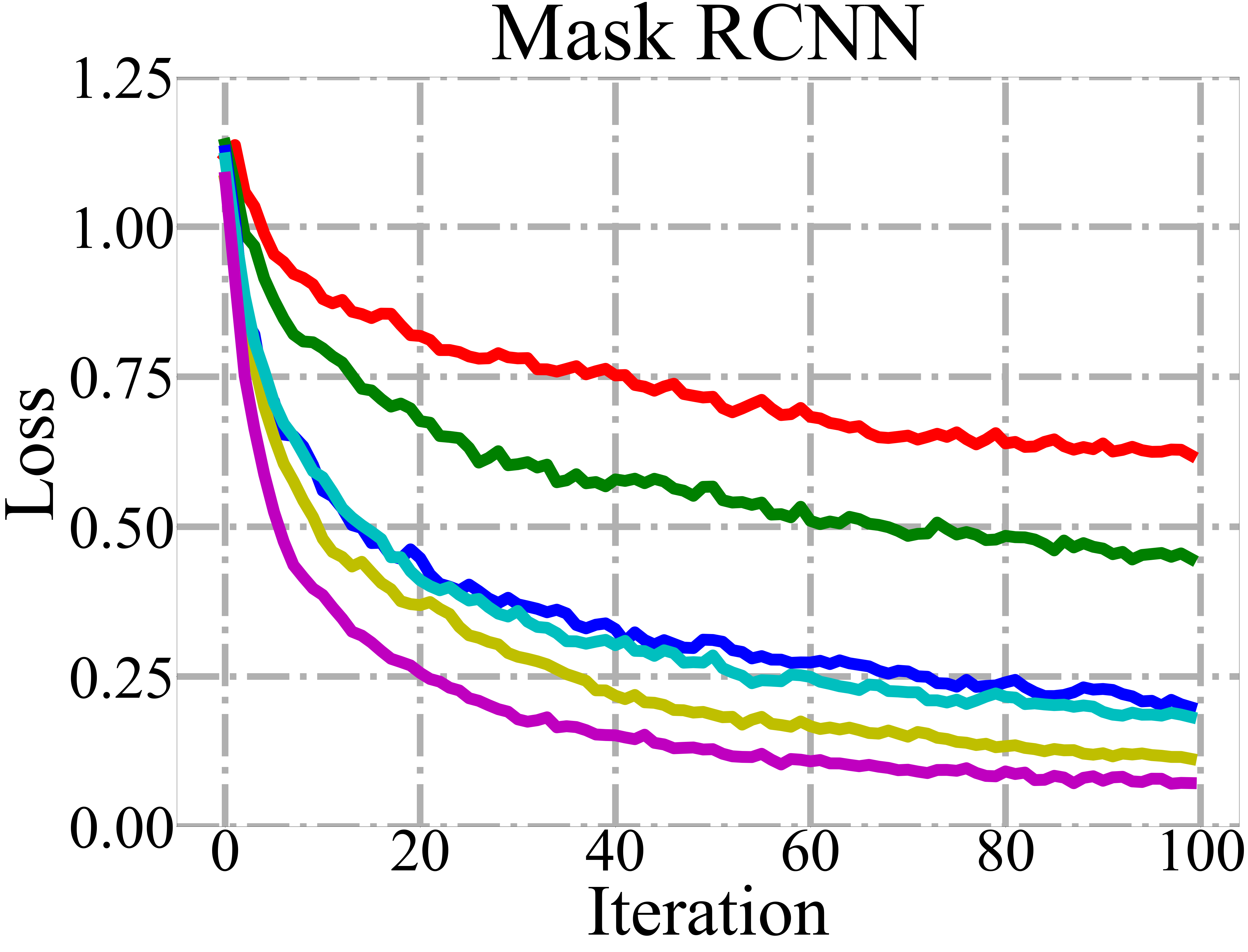}}
\end{minipage}
\begin{minipage}[]{.19\linewidth}
{               
\includegraphics[width=\linewidth]{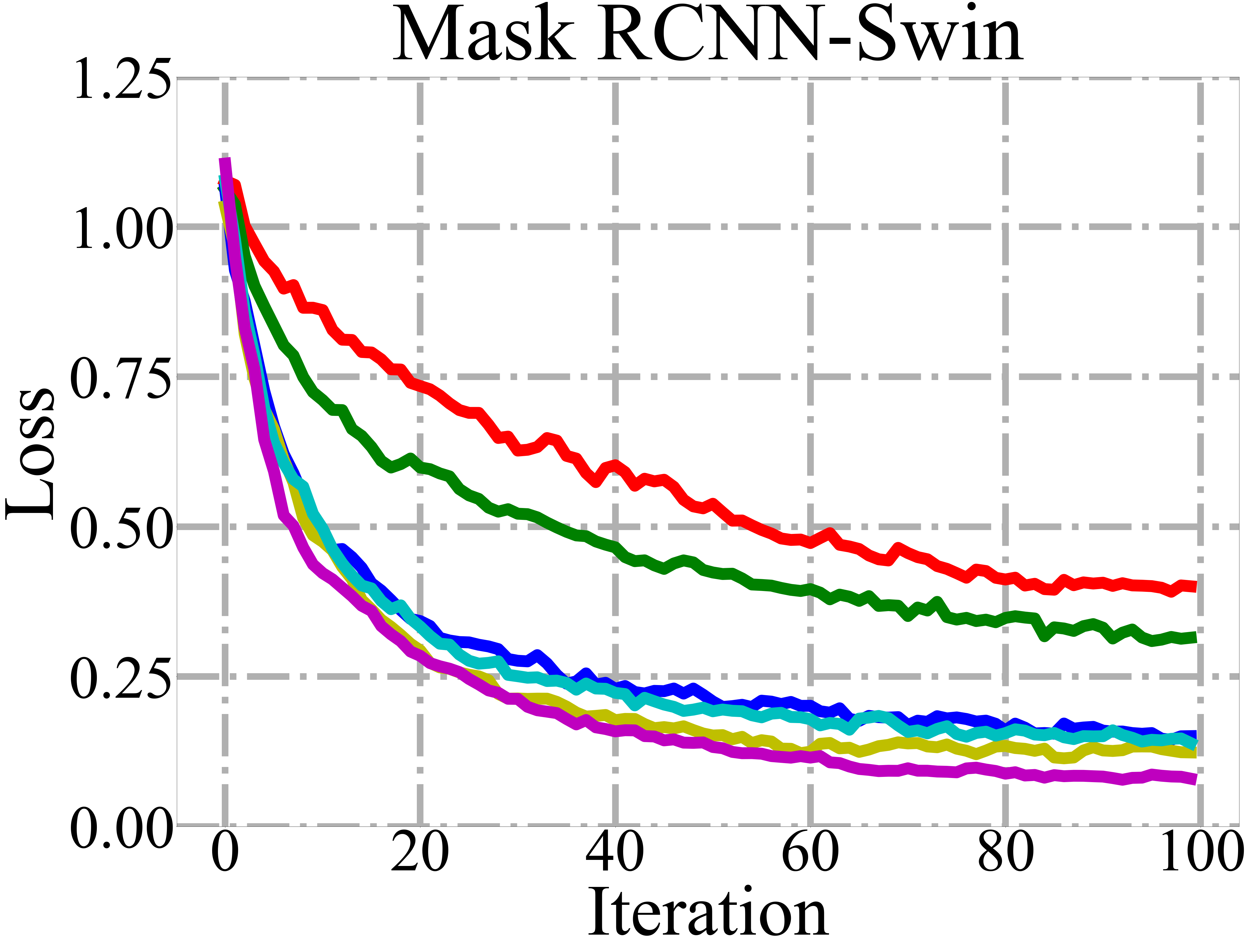}}
\end{minipage}
\hspace{0in}
\begin{minipage}[]{.19\linewidth}
{               
\includegraphics[width=\linewidth]{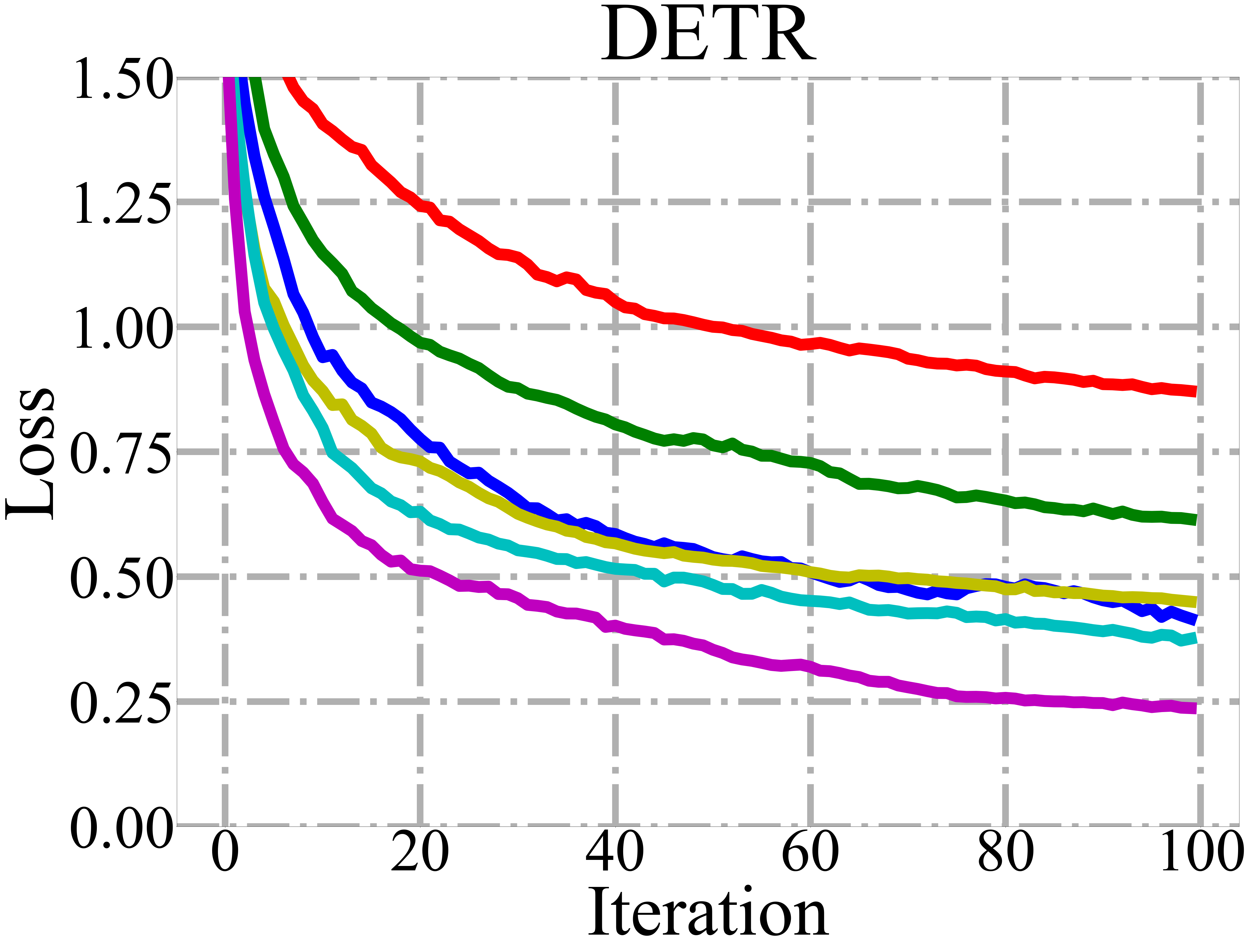}}
\end{minipage}
\hspace{0in}
\begin{minipage}[]{.19\linewidth}
{               
\includegraphics[width=\linewidth]{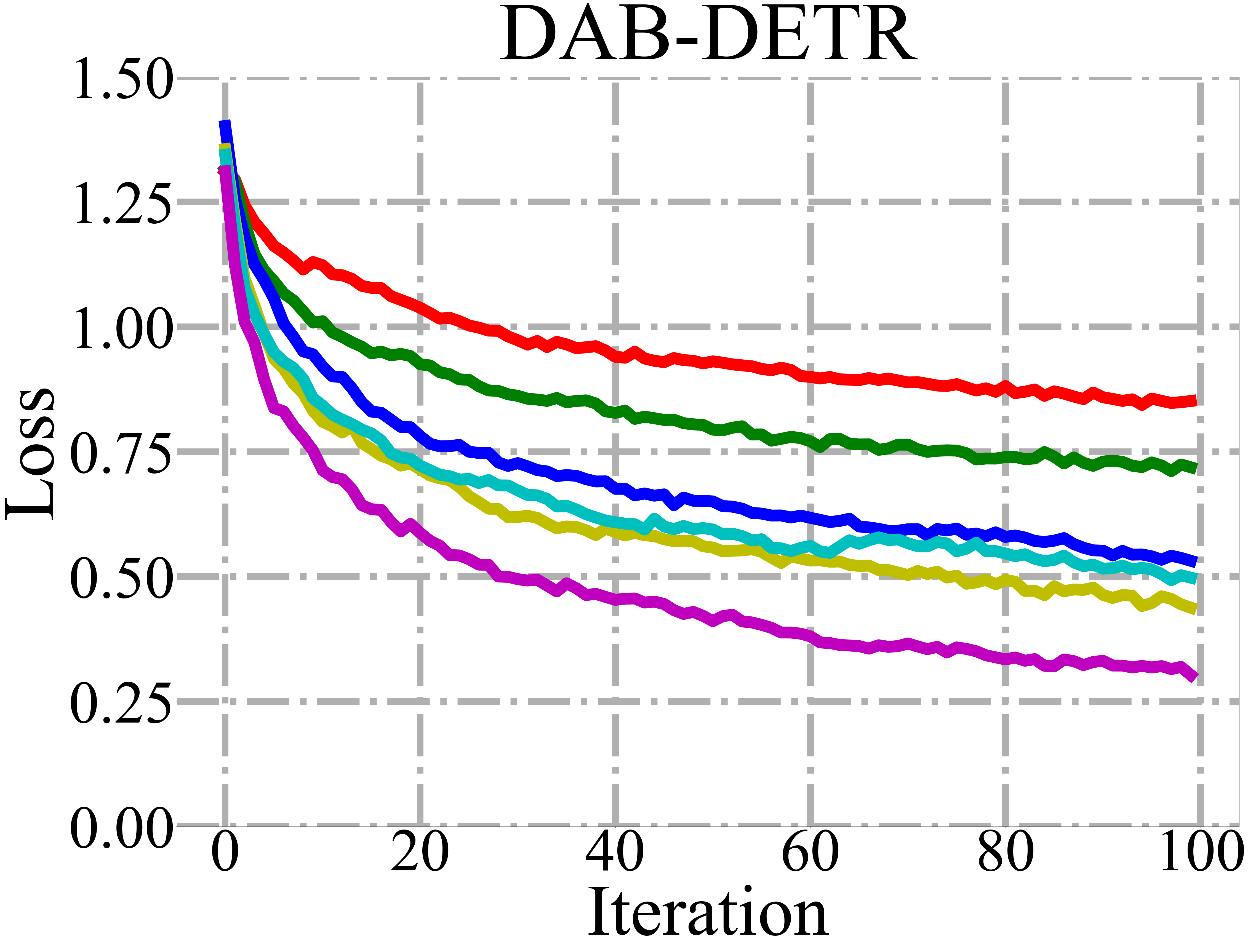}}
\end{minipage}
\hspace{0in}
\begin{minipage}[]{.19\linewidth}
{               
\includegraphics[width=\linewidth]{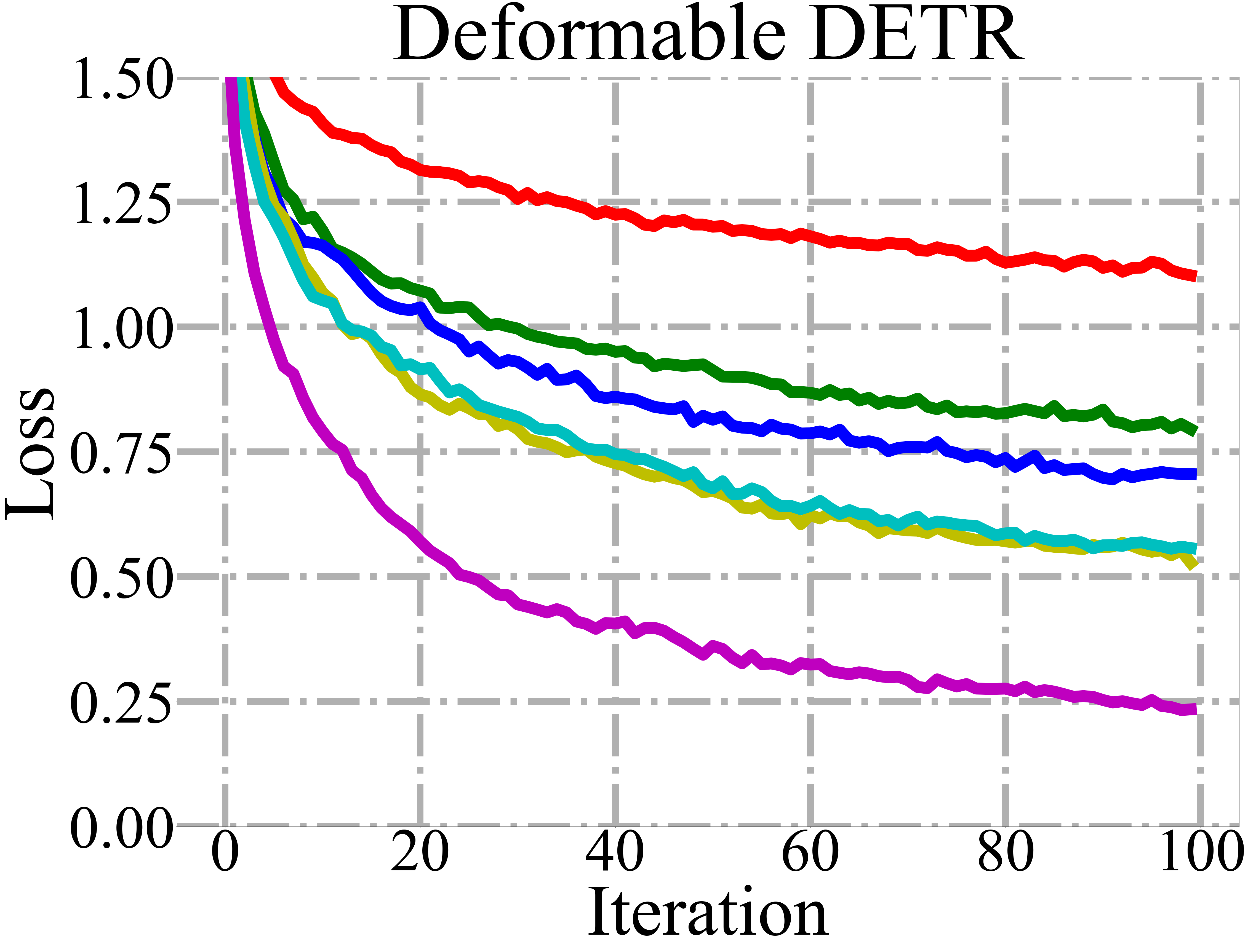}}
\end{minipage}
\hspace{0in}
\begin{minipage}[]{.19\linewidth}
{               
\includegraphics[width=\linewidth]{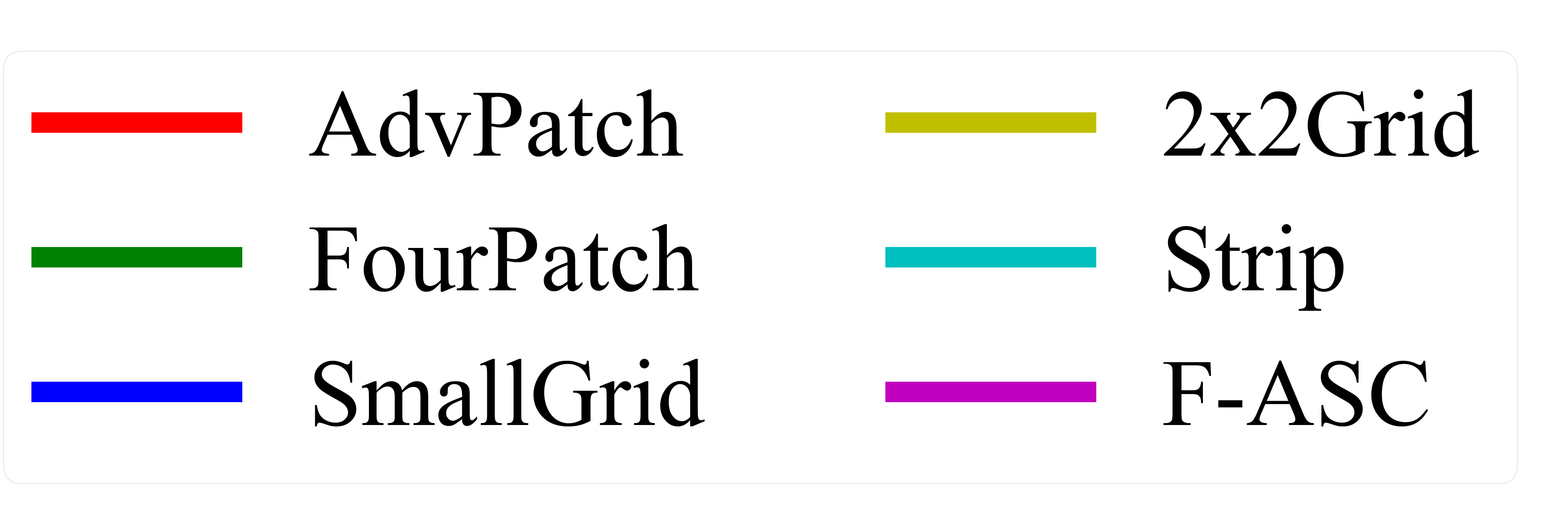}}
\end{minipage}
\caption{Convergence Curve on different types of models in the task of object detection in COCO.
}
\label{fig:curve}
\end{figure*}

\begin{table*}[!h]
\footnotesize
\centering
\caption{Successful Detection Rate (\%,$\downarrow$) for transfer-based black-box adversarial attacks across various detectors.}
\setlength{\tabcolsep}{0.8mm}{
\begin{tabular}{l|l|ccccccccccccccc}
\hline
  \multirow{2}{*}{\makecell{Ensemble\\models}}   &     & MRCN & \makecell{DAB-\\DETR} & \makecell{Retina\\Net} & \makecell{Corner\\Net}& \makecell{Cascade\\R-CNN} & SCNet & \makecell{Query\\Inst}  & TOOD & PAA  & \makecell{SABL\\RetinaNet} & DetectoRS & \makecell{Dynamic\\R-CNN} & \makecell{Auto\\Assign} & \makecell{Point\\Rend} & YOLACT \\  \cline{2-17} 
     & Clean          & 95.7 & 96.5     & 90.1      & 96.5      & 95.3 & 96.6  & 87        & 90.9 & 98.5 & 92.3 & 98.6      & 94.2          & 97.4       & 97.5        & 90.2   \\ \hline
\multirow{6}{*}{\makecell{\scriptsize{FRCN}\\\scriptsize{+YOLO}\\\scriptsize{+SSD}}}        & AdvPatch       & 90.1 & 94.4     & 83.2      & 94.1      & 91   & 93.5  & 82.1      & 82.5 & 95.9 & 89.1 & 97.6      & 88.4          & 95.9       & 96.1        & 86.1   \\
     & FourPatch      & 87.5 & 93.5     & 80.5      & 89.8      & 86.8 & 90.3  & 74.4      & 81.7 & 95.3 & 87.6 & 97.9      & 83.9          & 94         & 94.6        & 85     \\
     & SmallGrid      & 80.3 & 91.7     & 68.2      & 88.8      & 82.4 & 89.5  & 71.7      & 68.8 & 91.4 & 79.8 & 95.8      & 77.5          & 91.9       & 92.4        & 81     \\
     & 2$\times$2Grid & 75.4 & 88.4     & 66.7      & 81.8      & 77.3 & 86.3  & 66.6      & 63.6 & 88.4 & 77.8 & 96        & 73.5          & 89         & 91.7        & 75.5   \\
     & Strip          & 77.4 & 88.2     & 69        & \textbf{77.2}      & 74.7 & 85.9  & 62.4      & 71.4 & 90.5 & 79.5 & 95.9      & 69.6          & 89.9       & 90.1        & 76.1   \\
     & F-ASC          & \textbf{63.7} & \textbf{79.4}     & \textbf{52.8}      & 80.7      & \textbf{65.9} & \textbf{79.4}  & \textbf{53.5}      & \textbf{59}   & \textbf{82.4} & \textbf{70.9} & \textbf{94.4}      & \textbf{55.9}          & \textbf{84.7}       & \textbf{87.9}        & \textbf{72.7}   \\ \hline
\multirow{6}{*}{\makecell{\scriptsize{Swin-MRCN}\\\scriptsize{+YOLOX}\\\scriptsize{+DETR}}} & AdvPatch       & 95   & 94.8     & 88        & 94.4      & 94.5 & 95.3  & 81.7      & 86.1 & 97.1 & 90.1 & 98        & 93.1          & 96.1       & 96.6        & 88.2   \\
     & FourPatch      & 94.6 & 94.7     & 87.3      & 91.6      & 93.5 & 94.9  & 78.1      & 86.1 & 97.4 & 89   & 98.2      & 91.1          & 95         & 95.7        & 87.6   \\
     & SmallGrid      & 92.7 & 92       & 81.7      & 90.5      & 91.9 & 93.4  & 72.1      & 77.2 & 95.6 & 81.4 & 96.4      & 91.6          & 92.8       & 94.7        & 84.6   \\
     & 2$\times$2Grid & 89.9 & 88       & 80        & 84.4      & 88.1 & 91.3  & 68.1      & 74.5 & 94   & 81.5 & 96.6      & 86.3          & 91.4       & 93.2        & 79     \\
     & Strip          & 91.3 & 90       & 83        & 83.7      & 89.3 & 91.4  & 71.8      & 79.7 & 95.3 & 83.3 & 97        & 88.2          & 93.1       & 93.9        & 82.6   \\
     & F-ASC          & \textbf{85.7} & \textbf{82.3}     & \textbf{71.5}      & \textbf{83.6}      & \textbf{83.7} & \textbf{87.5}  & \textbf{62.8}      & \textbf{70.9} & \textbf{90.4} & \textbf{76.4} & \textbf{95.5}      & \textbf{81.5}          & \textbf{89}         & \textbf{92}          & \textbf{78.3} \\ \hline 
\end{tabular}
}
\label{tab:black}
\end{table*}

\textbf{Experiments on Auto-Driving Datasets.} Modern detectors have shown their power in scene analysis and are applied in many scenarios, such as auto-driving. We test the attack effectiveness of our method using auto-driving datasets of Cityscapes~\citep{Cordts2016Cityscapes} and BDD100K~\citep{bdd100k} toward object detection, specifically with object vanishing, to demonstrate that the fragility of DNN-based models should draw more caution to their application in safety-sensitive fields. We select 1000 objects of persons or vehicles, which are the most common objects in auto-driving, as the attack targets from each dataset respectively. As shown in~\cref{fig:auto-driving}, the pedestrian and bus in the front are cloaked by ASC, which uncovers the maximum vulnerability of modern detectors in the auto-driving scenarios. 

\cref{tab:cityscapes} and \cref{tab:bdd100k} shows the results of attacks in Cityscapes and BDD100K. ``F-ASC'' outperforms other prior patterns on all detectors and ``O-ASC'' outperforms the scattered $\ell_0$ attack methods on most of the detectors excluding SSD512. The trends of results are similar with those in COCO~\citep{lin2014microsoft} and suggests that the phenomena are statistically consistent. The results show that our method not only works effectively in COCO, but also demonstrates huge safety-risks to the auto-driving scenarios.

\subsection{Study on Efficiency}

In this section, we analyze the efficiency for different priors. We compare the convergence speed during the texture optimization of different prior patterns. We randomly select 100 images that can be successfully attacked by each fixed prior and draw the convergence curves of their average loss in the task of object vanishing with COCO in \cref{fig:curve}. We see the convergence speed of ``F-ASC'' is the fastest for 8 out of 9 detectors, except for Yolov3. The relative orders of convergence are similar to the attack effectiveness. The result indicates that the contour prior has stronger information and better efficiency when conducting attacks compared to other manually designed and meaningless priors. The faster convergence also implies the fact that modern detectors of different architecture are all sensitive to the area of object contour.

\begin{table*}[]
\centering
\caption{Successful Detection Rate (\%, $\downarrow$) in object mislabeling. }
\begin{tabular}{l|ccccccccc}
\hline
                & SSD512    & Yolov3    & YoloX     & FRCN      & MRCN      & MRCN-Swin & DETR      &  DAB-DETR     &    Def-DETR       \\ \hline
                \hline
Clean           & 94.6      & 91.1      &   99.4    & 97.7      & 98.3      &   99.5    & 99.4      &    98.3       &  96.9             \\
\hline
AdvPatch        & 4.4       & 9.8       &   31.7    & 82.0      & 85.3      &   73.0    & 91.8      &   73.6        &   66.1            \\
FourPatch       & 19.7      & 11.5      &  43.2     & 71.6      & 73.6      &   66.1    &  88.7     &   66.7        &   45.0            \\
SmallGrid       & 1.7       & 0.8       &   6.3     & 42.4      & 48.9      &  38.2     &  81.0     &  28.6         &    16.5           \\ 
2$\times$2Grid  & 0.5       & \bf{0.2}  &   4.6     & 24.7      & 27.5      &   31.7    &  70.8     &   16.2        &    8.0            \\
Strip           & 0.9       & 0.3       &   3.2     & 14.5      & 18.9      &   31.2    & 41.8      &    9.2        &    4.7            \\ 
0F-ASC           & \bf{0.2}  & 0.6       &  \bf{1.9} & \bf{7.4}  & \bf{8.9}  & \bf{16.1} &  \bf{19.8}&   \bf{5.0}    &    \bf{2.5}            \\\hline \hline 
C\&W-$\ell_0$   & \bf{0.0}  & 0.2       &   1.1     & 9.3       & 10.8      &  17.6     & 42.7      &  3.9          &     1.7         \\ 
PGD$_0$         &   0.5     &  0.5      &   11.2    &  7.1      &  9.8      & 16.6      &  14.2     &   7.0         &  3.1                 \\
O-ASC           & \bf{0.0}  & \bf{0.0}  &  \bf{0.0} & \bf{1.2}  & \bf{2.2}  &  \bf{3.9} & \bf{4.9}  &  \bf{0.9}     &   \bf{0.4}                \\ \hline
\end{tabular}
\label{tab:misclass}

\end{table*}

\begin{table*}[]
\centering
\caption{CIoU Distance ($\downarrow$)~\citep{zheng2020distance} in bounding box shift. }
\begin{tabular}{l|ccccccccc}
\hline
                & SSD512    & Yolov3    & YoloX     & FRCN      & MRCN      & MRCN-Swin  & DETR      &  DAB-DETR     &    Def-DETR      \\ \hline
\hline
Clean           & 0.8416    & 0.8239    &  0.9358   & 0.8712    & 0.8882    &  0.9125    & 0.9358    &  0.9231       &    0.9103               \\\hline
AdvPatch        &  0.0660   & 0.1595    &  0.3942   & 0.7045    & 0.7372    &  0.6929    &  0.8851   &  0.7317       &    0.6877               \\
FourPatch       &  0.2050   &  0.1539   &  0.4318   & 0.6336    & 0.6644    &   0.6822   & 0.8482    &  0.6544       &    0.4757               \\
SmallGrid       &  0.0299   & 0.0210    &  0.1123   & 0.4055    &  0.4576   & 0.4625     & 0.8193    &   0.3443      &    0.2280               \\ 
2$\times$2Grid  &\bf{0.0123}&\bf{0.0100}&  0.0862   & 0.2705    &  0.2907   &   0.3947   & 0.7441    &  0.2159       &    0.1073               \\
Strip           & 0.0234    & 0.0249    &  0.0730   & 0.1895    & 0.2108    &  0.4138    & 0.5044    & 0.1079        &    0.0633               \\
F-ASC           & 0.0170    & 0.0201    &\bf{0.0343}&\bf{0.0965}&\bf{0.1088}& \bf{0.2337}&\bf{0.2964}& \bf{0.0368}   &    \bf{0.0262}              \\  \hline\hline
C\&W-$\ell_0$   & 0.0134    & 0.0175    &  0.0277   & 0.1527    & 0.1633    &  0.2857    & 0.4700    &  0.0480       &  0.0385        \\ 
PGD$_0$         & 0.0089    & 0.0067    & 0.1793    & 0.0814    & 0.0922    &  0.1423    & 0.1783    & 0.0612        &    0.0232               \\
O-ASC           &\bf{0.0057}&\bf{0.0078}&\bf{0.0109}&\bf{0.0536}&\bf{0.0601}&\bf{0.1349} &\bf{0.1092}&  \bf{0.0122}   &   \bf{0.0093}                \\ \hline
\end{tabular}
\label{tab:shift}

\end{table*}

\subsection{Transfer-based Attack in Black-box Scenarios}

In order to demonstrate the benefits of a proper prior for the transfer-based black-box adversarial attack among different models, we use 1000 images from COCO and generate the adversarial examples for object vanishing on two sets of ensemble models respectively. One ensemble is Faster RCNN, SSD512 and YOLOv3, which are classic DNN-based detectors, while the other ensemble is Mask RCNN-Swin, YOLOX and DETR, which are with the latest and distinctive architectures. The transfer adversarial examples modify around 15\% pixels of the object area.

The results in~\cref{tab:black} show that ``F-ASC'' with object characteristics, has achieve the lowest SDRs on 15 black models, compared to other fixed prior patterns. Again, this proves that object contour prior shares the common characteristics of detectors with different architectures (
Mask R-CNN (MRCN)~\citep{he2017mask}, 
DAB-DETR~\citep{liu2022dabdetr}, 
RetinaNet~\citep{lin2017focal}, 
CornerNet~\citep{law2018cornernet}, 
Cascade R-CNN~\citep{cai2018cascade}, 
SCNet~\citep{liu2020scnet}, 
QueryInst~\citep{Fang_2021_ICCV},
TOOD~\citep{feng2021tood},
PAA~\citep{paa-eccv2020},
SABL RetinaNet~\citep{Wang_2020_ECCV},
DetectoRS~\citep{qiao2020detectors},
Dynamic R-CNN~\citep{DynamicRCNN},
AutoAssign~\citep{zhu2020autoassign},
PointRend~\citep{kirillov2019pointrend},
YOLACT~\citep{yolact-iccv2019}
), which can therefore degenerate their performance more effectively. The average drop of SDR among detectors with ``F-ASC'' compared to clean detection are 22.26\% and 12.41\%. The results show that the adversarial noises around object contour are more transferable, meaning that most modern detectors are adversarially sensitive and vulnerable to the adversarial noises around object contour. This further indicates a serious potential danger for real-world applications, \eg, autonomous driving systems.

\subsection{Additional Study}

In this section, we will deliver more additional experiments to further study the performance our proposed methods. Object detection involves object localization and classification simultaneously, which makes it a multi-task problem. Therefore, the attack objectives can also be multiple. Besides object vanishing which can be taken as a targeted attack, we can also generate adversarial examples to make the detector mislabel the objects or fail in localizing the objects. By carrying out these two tasks, we may further confirm the effectiveness of our method. Additionally, as we know that adversarial attack algorithms can have different behaviors on defense models, we also test our method on several defensive detectors. The experimental settings are consistent with those of the experiments on 1000 ``Person''s in COCO.

\textbf{Object Mislabeling.} In this task, we carry out an untargeted attack on the detectors by minimizing the classification confidence. 
We also use SDR as the metric with the additional condition that the classification of the object is correct. In general, \cref{tab:misclass} shows similar trends to object vanishing task. ``F-ASC'' leads other prior patterns almost all detectors, especially with large gaps on two-stage detectors and Detection Transformers. ``O-ASC'' further reduces the SDRs with sampling and outperforms all other traditional $\ell_0$ attack methods on each detector. The results demonstrate that ASC not only perturbs the prediction of object existence, but also can corrupt the classification.

\textbf{Bounding Box Shift.} In this task, we aim to minimize the Intersection over Union (IoU) as far as possible, which can be categorized as an attack on regression. We select Complete IoU (CIoU) ~\citep{zheng2020distance} as the metric. We only evaluate the positive CIoU and treat the cases with negative CIoU or no prediction as 0, since some examples may be entirely corrupted with no prediction given. \cref{tab:shift} shows the results of our experiments for bounding box shift with different methods. ``F-ASC'' can outperform all other prior patterns, except on SSD512 and Yolov3, where 2$\times$2Grid leads by a neck. After we have optimized the examples with high CIoUs, ``O-ASC'' has superior performance compared with other attackers on all the nine modern detectors. Though with looser constraints on noise distribution, C\&W-$\ell_0$ and PGD$_0$ still falls behind ``O-ASC''. This indicates that our method moves the bounding boxes farther away from the ground-truth label, suggesting that the contour pattern can offer more influence on the detector not only about the object existence, but also the object location.

\textbf{Defensive Detectors.} There are a variety of ways to have a defense model, including input transformation~\citep{guo2017countering}, adversarial training~\citep{pang2020bag}, and improved loss functions~\citep{li2021improving}. Notice the fact that we want to compare different attack methods on a defense model, using adversarial training is not reasonable enough. \revise{Therefore, we mainly consider the other two different types of defense and carry out the task of object vanishing on Faster RCNN with the experimental results shown in~\cref{tab:robust}. In consideration of the convenience for reading, we replicate the results of Faster R-CNN in~\cref{tab:disappear_person} as ``Vanilla'' here to make comparisons. It's straightforward to consider whether frequently-used smoothing and filtering techniques can influence the sparse noises generated by methods with better performance. We first adopt Gaussian Smoothing ($\sigma=4$) and Bilateral Filtering ($\sigma_x=\sigma_y=\sigma_r=1.5$) for denoising as simple defense strategies and use BPDA~\citep{obfuscated-gradients} to generate adversarial examples. From the results, we see that both techniques present some defense effectiveness for methods with more disperse noises. Still, the trends that F-ASC as fixed prior pattern outperforms other pre-defined patterns and O-ASC leads the performance over all baselines are consistent with previous results. Then, we take two methods that are more often used in adversarial defense.} We trained two robust detectors using Faster RCNN including the input manipulation of JPEG~\citep{guo2017countering} and an improved loss function of Probabilistically Compact Loss (PC Loss)~\citep{li2021improving}. 
The input compression by JPEG is to neutralize the influence of adversarial noises and the usage of PC Loss instead of Cross Entropy Loss is mainly to enlarge the gaps of classification probabilities and therefore strengthen the robustness. Though the defense sacrifices the performance on clean detection a little, the improvements in adversarial robustness are significant.
For JPEG Compression, we adopt BPDA~\citep{obfuscated-gradients} with all methods by using differentiable computation to approximate the compression, and as a result, ``F-ASC'' achieves the lowest SDR compared to the other five patterns while ``O-ASC'' leads to further improvement and outperforms C\&W-$\ell_0$ and PGD$_0$.
As for PC Loss, we take the same attack method, and see that our methods still have better attack effectiveness.
This result suggests that the proposed contour-based sparse attack works well and can be extended to the defense models.

\begin{table}[!t]
\centering
\scriptsize
\caption{Successful Detection Rate (\%, $\downarrow$) with defenses on FRCN. 
}
\setlength{\tabcolsep}{0.9mm}{
\begin{tabular}{l|c|cccccc|ccc}
\hline
      &Clean & AP   & FP & SG & 2$\times$2G & Strip & F-ASC & C\&W  & PGD$_0$ & O-ASC\\ \hline
\hline
Vanilla     & 98.3 & 85.0 & 76.4 & 46.9 & 28.3 & 16.8 & \bf{9.7} & 13.1 & 8.8 & \textbf{2.0} \\ \hline
\textcolor{black}{Gaussian}    & \textcolor{black}{98.3} &	\textcolor{black}{85.6}&	\textcolor{black}{74.7}&	\textcolor{black}{46.3}&	\textcolor{black}{27.6}&	\textcolor{black}{17.7}&	\textcolor{black}{\textbf{10.1}}&	\textcolor{black}{13.7}&	\textcolor{black}{17.4}&	\textcolor{black}{\textbf{4.3}}\\\hline
    \textcolor{black}{Bilateral} & \textcolor{black}{97.4}&	\textcolor{black}{82.8}&	\textcolor{black}{70.5}&	\textcolor{black}{38.6}&	\textcolor{black}{23.6}&	\textcolor{black}{12.5}&	\textcolor{black}{\textbf{10.2}}&	\textcolor{black}{16.9}&	\textcolor{black}{22.3}&	\textcolor{black}{\textbf{5.6}}\\\hline
JPEG       & 97.9 & 88.0 & 79.1 & 57.0 & 42.1 & 27.4 & \bf{20.1} & 32.5 & 26.0 & \textbf{7.9} \\ \hline
PCLoss   & 97.6 & 84.6 & 81.0 & 58.8 & 45.1 & 42.8 & \bf{28.5}  & 22.4 & 18.0 & \textbf{15.7} \\ \hline
\end{tabular}}
\label{tab:robust}
\end{table}

\section{Conclusion}

In this paper, we propose a novel prior-guided adversarial attack on object detection, which provides a unifying view of most current sparse adversarial attacks under the Bayesian framework. To address the limitations of existing methods, especially those manually designed, we introduce a more informative prior of object contours, which are the common characteristics for various object detectors. With the prior contour appropriately harnessed, we optimize the textures, leading to a more effective and efficient adversarial attacker, and sample the pixels around the contour to further improve the attack performance and outperform the traditional $\ell_0$ methods with scattered perturbations. Extensive experiments demonstrate the superior performance compared to other methods on both COCO and the auto-driving datasets. Our work concludes with the phenomenon that object contour serves as a common weakness for various modern detectors and they are more sensitive to adversarial noises in that area. This further raises concerns over the applications of these DNN-based object detectors in safety-critical systems, \eg, autonomous driving systems. 


\section*{Acknowledgments}

This work was supported by the National Key Research and Development Program of China (2020AAA0106000, 2020AAA0104304, 2020AAA0106302), NSFC Projects (Nos. 62061136001, 62076147, 62071292, 61771303, U19B2034, U1811461, U19A2081), Beijing NSF Project (No. JQ19016), STCSM Project (No. 18DZ2270700), Tsinghua-Alibaba Joint Research Program, Tsinghua Institute for Guo Qiang, Tsinghua-OPPO Joint Research Center for Future Terminal Technology, and the High Performance Computing Center, Tsinghua University.

\bibliographystyle{model2-names}
\bibliography{egbib}

\end{document}